\documentclass[utf8]{FrontiersinVancouver}

\usepackage{url,hyperref,microtype,subcaption}

\usepackage[onehalfspacing]{setspace}

\usepackage{multirow} 
\usepackage{longtable} 

\usepackage{color,soul}

\usepackage{pdflscape} %
\usepackage{multicol}

\graphicspath{{./images/}}

\def\keyFont{\fontsize{8}{11}\helveticabold }
\def\firstAuthorLast{Abbas {et~al.}} 
\def\Authors{Saram Abbas\,$^{1,*}$, Rishad Shafik\,$^{1}$, Naeem Soomro\,$^{2}$, Rakesh Heer\,$^{3,4}$ and Kabita Adhikari\,$^{1 *}$}
\def\Address{$^{1}$School of Engineering, Newcastle University, UK \\
$^{2}$Department of Urology, Freeman Hospital, Newcastle upon Tyne, UK. \\
$^{3}$Division of Surgery, Imperial College London, London, UK\\ 
$^{4}$Newcastle University Centre for Care, Newcastle upon Tyne, UK}

\def\corrAuthor{Saram Abbas}

\def\corrEmail{s.abbas11@newcastle.ac.uk}

\begin{document}
\onecolumn
\firstpage{1}

\title[Predicting NMIBC Recurrence using AI]{Reviewing AI's Role in Predicting Recurrence of Non-Muscle-Invasive Bladder Cancer} 

\author[\firstAuthorLast ]{\Authors} 
\address{} 
\correspondance{} 

\extraAuth{}%

\maketitle

\begin{abstract}

\section{}
Notorious for its 70-80\% recurrence rate, Non-muscle-invasive Bladder Cancer (NMIBC) imposes a significant human burden and is one of the costliest cancers to manage. Current prediction tools for NMIBC recurrence rely on scoring systems that often overestimate risk and have poor accuracy.  This is where Machine learning (ML) and artificial intelligence are tranforming oncological urology by leveraging molecular and clinical data. This comprehensive review  critically analyses ML-based frameworks for predicting NMIBC recurrence, focusing on their statistical robustness and algorithmic efficacy. We meticulously examine the strengths and weaknesses of each study, by focusing on various data modalities, and ML models, highlighting their performance alongside inherent limitations. A diverse array of ML algorithms, including neural networks, deep learning, and random forests, demonstrates immense potential in enhancing predictive accuracy by leveraging multimodal data spanning radiomics, clinical, histopathological, and genomic domains. However, the path to widespread adoption faces challenges concerning the generalizability, interpretability and explainaibility, emphasising the need for collaborative efforts, robust datasets, and the incorporation of cost-effectiveness. Our detailed categorization and in-depth analysis illuminate the nuances, complexities, and contexts that influence real-world advancement and adoption of these AI-driven techniques in precision oncology. This rigorous analysis equips researchers with a deeper understanding of the intricacies of the ML algorithms employed. Actionable insights are provided for researchers aiming to refine algorithms, optimise multimodal data utilisation, and bridge the gap between predictive accuracy and clinical utility. This review serves as a roadmap to advance real-world AI applications in oncological care.
\tiny
 \keyFont{ \section{Keywords:}   Artificial intelligence, Non-muscle-invasive bladder cancer, NMIBC, Machine learning, Recurrence, Prediction } 
\end{abstract}

\section{Introduction}

Bladder cancer continues to be a significant health concern. Particularly in the UK, where it stands as the 11th most common cancer, necessitating efficacious diagnostic and management strategies to curtail its impact \cite{BladderCancerStatistics2015}. With 28 new reported cases and 15 deaths daily, the disease's impact is undeniable \cite{BladderCancerStatistics2015}. A disease of heterogeneous nature, bladder cancer is primarily categorised into two main types: non-muscle-invasive bladder cancer (NMIBC) and muscle-invasive bladder cancer (MIBC). Among them, NMIBC is more common, yet carries a risk of escalating into MIBC if left untreated or poorly managed. About 45\% of untreated high-grade NMIBC cases escalate into MIBC \cite{vandenboschLongtermCancerspecificSurvival2011,babjukEAUGuidelinesNon2017,hallGuidelineManagementNonmuscle2007}. An alarming feature of NMIBC is its high recurrence rate post-treatment, reported to be 70-80\% \cite{witjesEAUGuidelinesMuscleinvasive2014}, requiring frequent monitoring and interventions.

Due to its high prevalence and recurrence rate, bladder cancer happens to be one of the costliest cancers to manage – it cost the EU approximately €4.9 billion in 2012 to treat bladder cancer \cite{mossanenBurdenBladderCancer2014,lealEconomicBurdenBladder2016a}. The high cost of managing NMIBC is due in part to the expensive diagnostic procedures and follow-up care required for patients with recurrent NMIBC.  For example, cystoscopy, the gold standard for monitoring, is expensive (£240-£2000 per visit) and invasive, contributing significantly to overall costs and involving indirect costs from lost productivity and risks like urinary tract infections \cite{tanMixedMethodsApproach2019,loStrategiesPreventCatheterAssociated2014}. From 2015-2022, flexible cystoscopy cost the NHS over £810 million {in total } (inflation-adjusted) \cite{NHSEnglandNational2021, NHSEngland20202020, NHSEngland20192019, NHSEngland20182018, NHSReferenceCosts2015,InflationCalculator2023}. Guidelines recommend cystoscopy every 3-6 months for the first 2 years, then annually, leading to significant cumulative costs. The average 3-year cost per NMIBC patient in the UK was estimated at £8735,with annual costs ranging from £1218 for grade 1 recurrence cases to £3957 for grade 3 \cite{coxImpactsBladderCancer2019}. These escalating costs juxtapose the financial burden of NMIBC with the pressing need for precise diagnostic techniques and reliable predictive tools that can reduce the frequency of costly cystoscopic follow-ups.

Traditionally, prognostication and risk assessment in bladder cancer have relied on tools like the American Joint Committee on Cancer TNM staging system \cite{kandoriUpdatedPointsTNM2019,edgeAmericanJointCommittee2010}. The TNM system categorizes bladder cancer based on tumour size and invasion (T), lymph node involvement (N), and distant metastasis (M). While validated and widely used, these systems lack the comprehensive integration of factors needed for precise prognostication \cite{changDiagnosisTreatmentNonMuscle2016}. They do not encompass the full scope of factors necessary for precise prognostication. They seem to fall short when compared to predictive models that incorporate numerous clinical variables \cite{changDiagnosisTreatmentNonMuscle2016}. Additionally, their design does not readily permit the integration of novel information such as molecular markers or complex bioinformatics data, which are becoming increasingly relevant in the era of personalised medicine \cite{hensleyContemporaryStagingMuscleInvasive2022}. A multimodal approach is recommended which combines histopathological markers with the results of imaging studies. 

Despite numerous published studies on NMIBC recurrence prediction models, their adoption in clinical practice remains surprisingly low. Reasons for this can range from the lack of demonstrated improvement in clinical decision-making upon external validation to the logistical difficulty of integrating these prediction tools into electronic medical records to be readily available to the physician at the point of care. This systematic review addresses these gaps by analyzing AI-based prediction models for bladder cancer recurrence, emphasizing the need for better validation and integration into clinical workflows.

The aim of this systematic review is to evaluate the performance and utility of AI-based predictive models in NMIBC recurrence. The research question, framed using the PICOS criteria, is as follows: \textit{How do AI-based predictive models perform compared to traditional methods in forecasting recurrence in NMIBC patients?} Specifically:
\begin{itemize}
    \item \textbf{Population:} Patients diagnosed with NMIBC.
    \item \textbf{Intervention:} AI-based predictive models.
    \item \textbf{Comparison:} Traditional statistical or clinical models.
    \item \textbf{Outcome:} Predictive performance metrics (e.g., accuracy, discrimination, AUC) and clinical utility.
    \item \textbf{Study Design:} Observational, retrospective, and prospective studies.
\end{itemize}

By synthesizing the findings of existing studies, this review aims to identify gaps in current methodologies, provide insights into effective AI-based approaches, and offer guidance for future research to enhance personalized and efficient bladder cancer management.

\subsection{Non-muscle invasive Bladder Cancer}

\begin{figure*}[!t]
	
	\centering
	\includegraphics[width=0.55\textwidth]{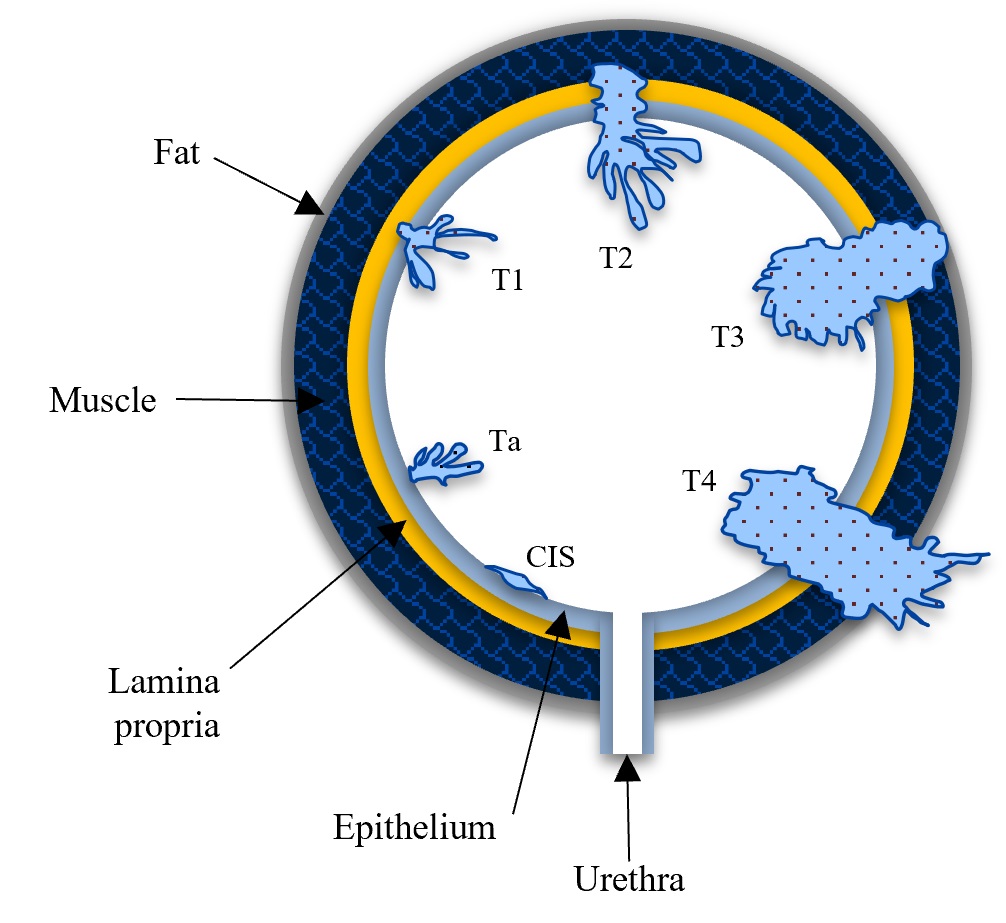}
	\caption[short]{Stages of Tumour metastasis illustrated in AJCC TNM staging system.  Carcinoma-in-situ (CIS), Ta and T1 are non-muscle invasive stages and T2 - T4 are muscle invasive stages. CIS: primary stage where tumour is confined to inner bladder lining. Ta: tumour limited to epithelium. T1: tumour reaches the lamina propria. Stage III (T2): tumour invades into bladder wall muscle. T3: tumour spreads to the fat around the bladder. Stage IV (T4): tumour spreads to nearby pelvic organs/tissues.}
	\label{figure_1}
\end{figure*}

Bladder cancer is a disease characterized by the uncontrolled growth of abnormal cells in the lining of the bladder, a hollow organ in the lower abdomen responsible for storing urine. Non-muscle-invasive bladder cancer (NMIBC) is a subtype of bladder cancer that hasn't penetrated the muscular wall of the bladder. This type of bladder cancer, often found early, comprises about 75\% of all bladder cancer cases \cite{kamatBladderCancer2016}.

Bladder cancer staging follows the TNM system, detailing cancer progression within or beyond the bladder (see Figure~\ref{figure_1} for TNM staging). {} Early stages (CIS, Ta, T1) are Non-Muscle-Invasive, with the tumour confined to the bladder's surface or connective tissue yet not penetrating the deeper muscle layers. These stages can, however, progress to muscle-invasive (MIBC) stages. NMIBC carries a lower risk of metastasis compared to MIBC but has a high recurrence rate \cite{kamatBladderCancer2016}. NMIBC has  a tendency to often recur. These recurrences can be either at the same stage as the initial tumour or at a more advanced stage \cite{clarkClinicalPracticeGuidelines2013}. Treatment primarily includes transurethral resection of bladder tumour (TURBT) to remove cancerous cells \cite{richardsImportanceTransurethralResection2014}. Depending on the risk of recurrence and progression, intravesical therapy, where medication is directly instilled into the bladder, may be applied post-TURBT. This can include chemotherapy agents, such as mitomycin C, or immunotherapy with Bacillus Calmette-Guerin. 

\begin{figure*}[!t]
	\centering
	\includegraphics[scale=0.3]{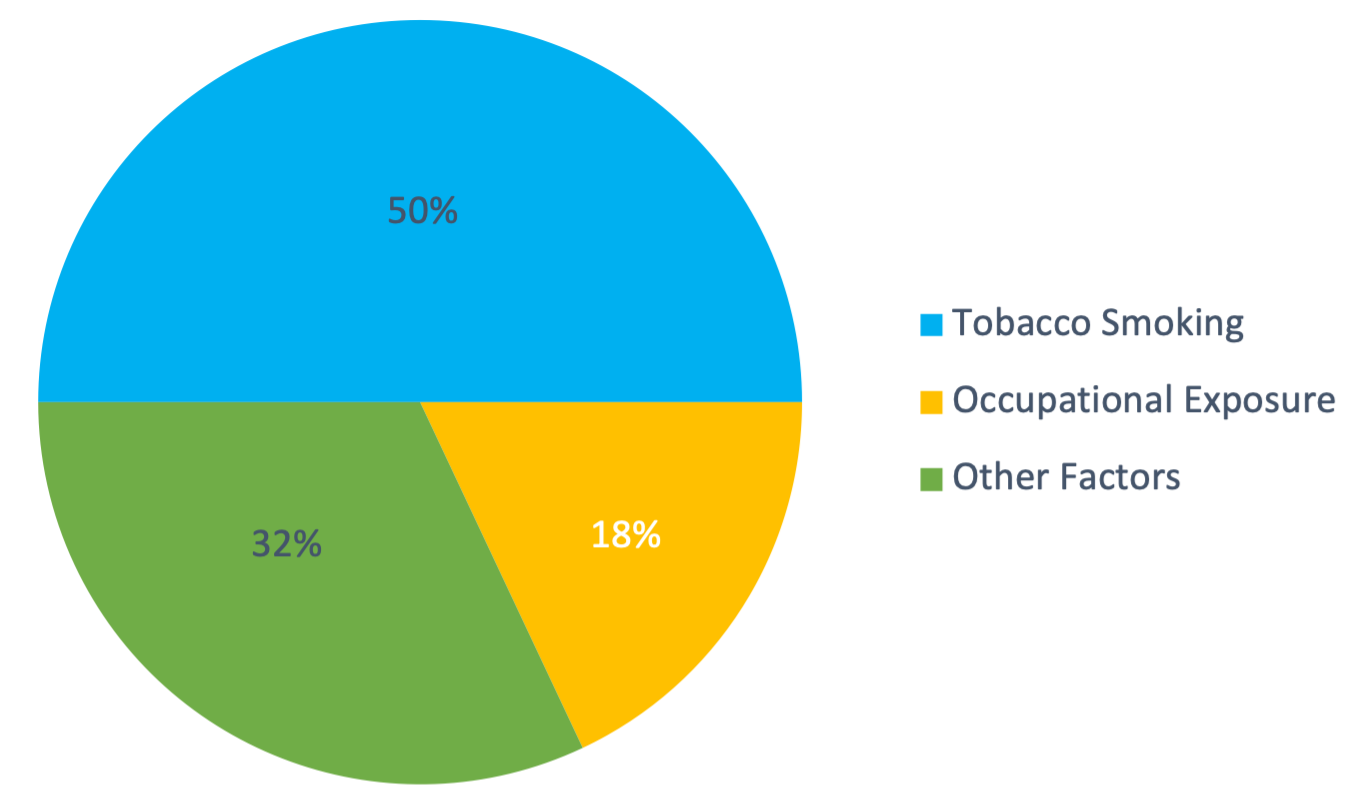}
	\caption[short]{Distribution of major risk factors for bladder cancer. Tobacco smoking accounts for approximately 50\% of cases, occupational exposures contribute to 18\%, and other factors (including exposure to arsenic, chronic bladder inflammation, previous radiation or chemotherapy, diet, and genetic predisposition) comprise the remaining 32\%. Data synthesized from multiple sources \cite{burgerEpidemiologyRiskFactors2013, letasiovaBladderCancerReview2012, farlingBladderCancerRisk2017, chenBladderCancerScreening2005, cumberbatchEpidemiologyBladderCancer2018}.}
	\label{risk_factors}
\end{figure*}

Bladder cancer is associated with numerous risk factors, the most significant being tobacco smoking, which accounts for about half of all cases \cite{burgerEpidemiologyRiskFactors2013}. Figure ~\ref{risk_factors} highlights the overwhelming effect of smoking over other factors. Other factors include occupational exposures, chronic bladder inflammation, and genetic predisposition, which contribute to the complexity of predicting the disease's occurrence and progression \cite{chenBladderCancerScreening2005, letasiovaBladderCancerReview2012, cumberbatchEpidemiologyBladderCancer2018}. Additionally, gender disparities exist, with men being about three to four times more likely to develop bladder cancer than women \cite{brayGlobalCancerStatistics2018}. Age, diet high in processed meat, and tumour characteristics like grade and size also play crucial roles in predicting the likelihood of recurrence in NMIBC \cite{burgerEpidemiologyRiskFactors2013, letasiovaBladderCancerReview2012, farlingBladderCancerRisk2017}. Other predictive factors include genetic alterations or mutations, the presence of bladder cancer markers in the urine, and findings from imaging studies and cystoscopy \cite{audenetEvolutionBladderCancer2018}. This variability in risk factors highlights the challenges in developing personalized care plans, underscoring the need for sophisticated predictive tools.

Creating personalised care plans for bladder cancer is challenging due to its highly varied nature. Each case differs in terms of tumour biology, stage, grade, and genetic mutations, which means a treatment effective for one patient may not work for another. Despite advances in bladder cancer genetics, the lack of reliable predictive biomarkers underscores the need for more advanced tools. In the following sections, we explore how Artificial Intelligence and Machine Learning (ML)-based approaches, are poised to fill this gap by offering more precise and individualized predictions, ultimately improving patient outcomes and treatment strategies.

\subsection{Current Methods and the Need for Innovation}

The European Organisation for Research and Treatment of Cancer (EORTC) and The Club Urológico Español de Tratamiento Oncológico (CUETO), are two popular clinical urology research organisations, that developed the tools for recurrence prediction used currently. EORTC developed the EORTC Risk Tables \cite{EORTCRiskTables2023}, and CUETO introduced the CUETO Scoring system. These tables use a variety of factors, such as the number and size of tumours, prior recurrence rates, T-stage, and grade to calculate risk scores \cite{seoEfficacyEORTCScoring2010, PredictingDiseaseRecurrence2023}. Vedder et al. \cite{vedderRiskPredictionScores2014} conducted a study which revealed that both EORTC and CUETO's risk scores were not accurate in predicting recurrence (found c-indices of 0.55-0.61, where 0.5 is random guess). Other studies have also concluded that CUETO and EORTC presented poor discriminative value in predicting clinical events. These models overestimated the risk, especially in highest-risk patients \cite{krajewskiAccuracyCUETOEORTC2022,fujiiPredictionModelsProgression2018,xylinasAccuracyEORTCRisk2013}. This has significant implications for both physicians and patients alike. 

Poor and unreliable recurrence prediction has far-reaching consequences. Unreliable predictions of recurrence can trigger unnecessary invasive procedures like repeated cystoscopies or biopsies, overburdening healthcare providers and depleting valuable resources \cite{bornReducingHarmOveruse2022}. This, in turn, escalates the financial costs borne by the NHS \cite{WarningCutsNHS2021,greenfieldNHSWieldsAxe2018}, diverting funds that could otherwise be allocated to patient care and research endeavours. For patients, the repercussions of unreliable recurrence prediction are substantial, entailing risks and complications. False-positive predictions cause unnecessary anxiety and subject patients to additional tests and treatments with inherent risks. Conversely, false-negative predictions can delay or miss disease progression, hindering timely intervention and adversely affecting patient health \cite{monteiroICUDSIUInternationalConsultation2019,kimTransurethralResectionBladder2020}.

Traditional mathematical and statistical tools use a limited set of variables. These tools assume straightforward, proportional relationships between patient variables like tumour size, number of tumours, prior recurrence, and T-stage with outcomes such as recurrence and progression. By assigning linear scores based on cancer grade (1, 2, or 3), they overlook the nuanced severity differences, such as the greater jump from Grade 2 to 3 compared to Grade 1 to 2. In reality, cancer prognosis is shaped by complex, non-linear interactions involving molecular markers, genetic factors, and evolving medical data—factors that static, linear models cannot adequately capture. These traditional methods also struggle with missing data and incorporating recent advancements. Emerging approaches aim to enhance tumour classification, discover novel biomarkers, and improve predictions for bladder cancer metastasis, while other promising areas include body composition analysis and biomarkers beyond traditional assessments \cite{zhuTraditionalClassificationNovel2020,castanedaIdentifyingNovelBiomarkers2023, zhangEmergingBiomarkersPredicting2021}. Additionally, novel prognostic models based on gene signatures show potential \cite{huangBodyCompositionPredictor2023}. However, these innovations cannot be integrated into traditional tools highlighting the urgent need for more sophisticated models that offer reliable, dynamic predictions for NMIBC recurrence.

\subsection{AI as a game-changer in NMIBC Recurrence Prediction}
Artificial Intelligences (AI) has emerged as a powerful tool in the medical field, with machine learning (ML) at its forefront, particularly in tasks related to cancer prognosis and recurrence prediction. AI, a broad category of computational methodologies designed to emulate human cognitive functions, has been increasingly deployed in the medical field. ML algorithms can be trained to learn from existing data, adjusting their mathematical parameters to predict outcomes with high accuracy. In the context of bladder cancer, algorithms like support vector machines (SVM), random forest (RF), artificial neural networks (ANN), and deep learning (DL) have been used to design {models that } enhance the prediction of cancer recurrence \cite{xuPredictiveNomogramIndividualized2019,shkolyarAugmentedBladderTumor2019,tokuyamaPredictionNonmuscleInvasive2022,pantazopoulosBACKPROPAGATIONNEURAL1998,shaoMetaboliteMarkerDiscovery2017,cattoArtificialIntelligencePredicting2003a}.

These models leverage a diverse range of markers — radiomic, clinical, pathological, and genomic — to build comprehensive and nuanced predictive models \cite{lambinRadiomicsBridgeMedical2017,zhengAccurateDiagnosisSurvival2022,bychkovDeepLearningBased2018,yuanDeepGeneAdvancedCancer2016,mobadersanyPredictingCancerOutcomes2018}. They aim to improve risk stratification, anticipate recurrence, and optimise treatment planning, opening doors to more personalised patient management. Nonetheless, while advancements in AI and ML hold considerable promise, there is still much to learn, explore, and validate before these technologies can fully realise their potential in bladder cancer management.

\begin{figure*}
	\centering
	\includegraphics[width=1\textwidth]{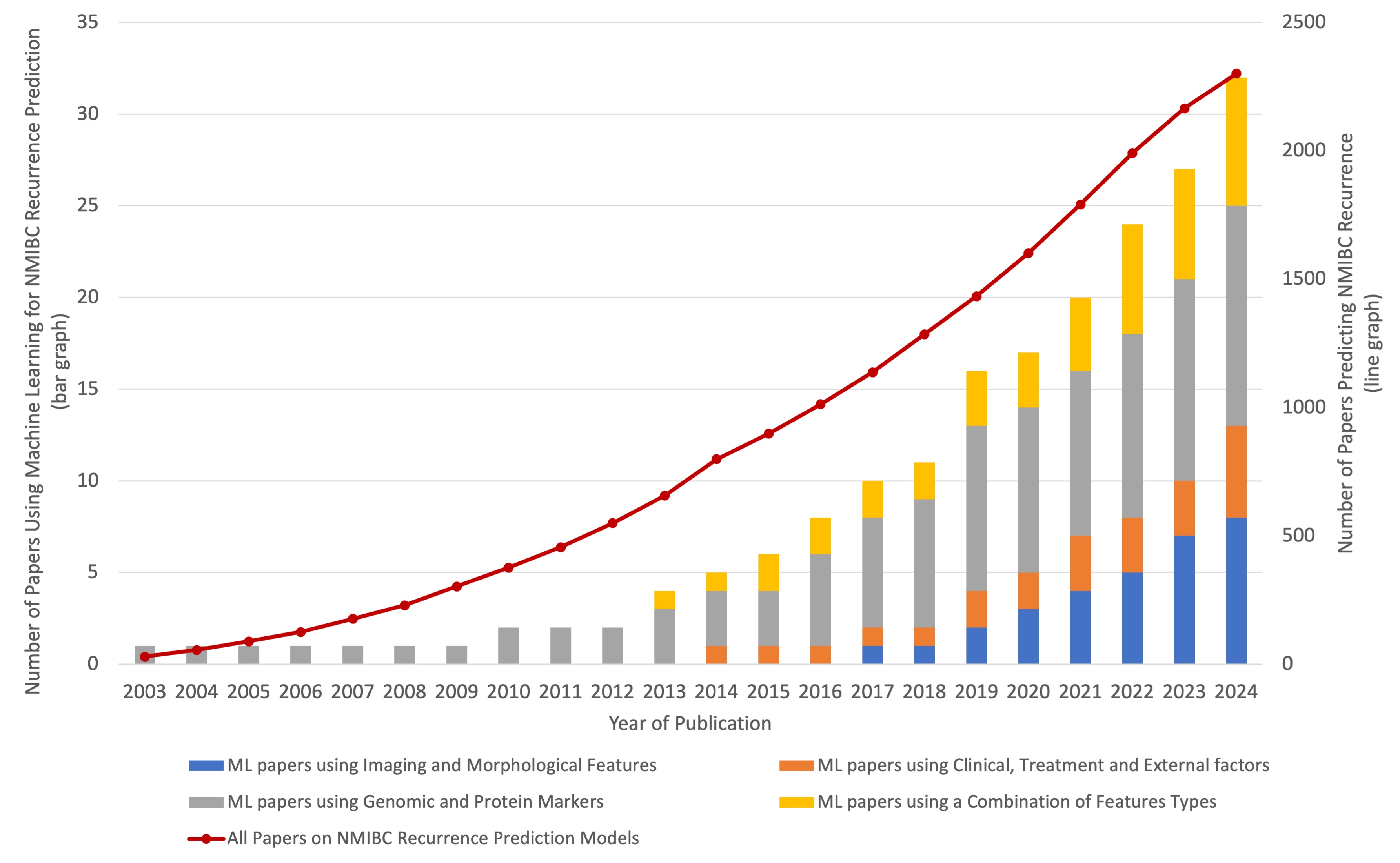}
	\caption{This graph shows the overall growth in bladder cancer recurrence studies and the exponential rise in ML-based approaches. While the total number of bladder cancer studies has increased linearly over the past two decades, the adoption of ML methods has grown at a much faster, exponential rate. This trend suggests that ML approaches are becoming increasingly relevant in the field, with future research likely to continue emphasizing advanced computational methods.}
	\label{MLTrends}
\end{figure*}

Research in AI techniques to predict NMIBC recurrence is accelerating rapidly. As demonstrated in Figure~\ref{MLTrends}, the number of studies focusing on bladder cancer recurrence has shown a steady, linear increase over the past two decades (depicted by a line graph). However, the adoption of ML approaches has grown at a much faster, exponential rate (depicted by the bar graph). This surge in ML research highlights the recognition of its potential to improve predictive accuracy and model complex, non-linear relationships that traditional statistical methods often struggle to capture. With increasing accessibility and advancements in computational power, the relevance of ML approaches in bladder cancer research is expected to continue expanding.

An emerging trend in the development of ML models for NMIBC prediction is the integration of multiple feature types, such as clinical, genomic, and imaging data. Single-feature models often fail to account for the multifaceted nature of cancer progression, whereas combining multiple data sources allows for more comprehensive models and improved predictive performance. This shift can be attributed to advancements in data integration techniques, the growing availability of multi-modal datasets and ease of access of powerful GPUs, which enable researchers to leverage richer and more diverse data for more accurate and clinically relevant predictions.

\begin{figure*}
	\centering
	\includegraphics[width=1\textwidth]{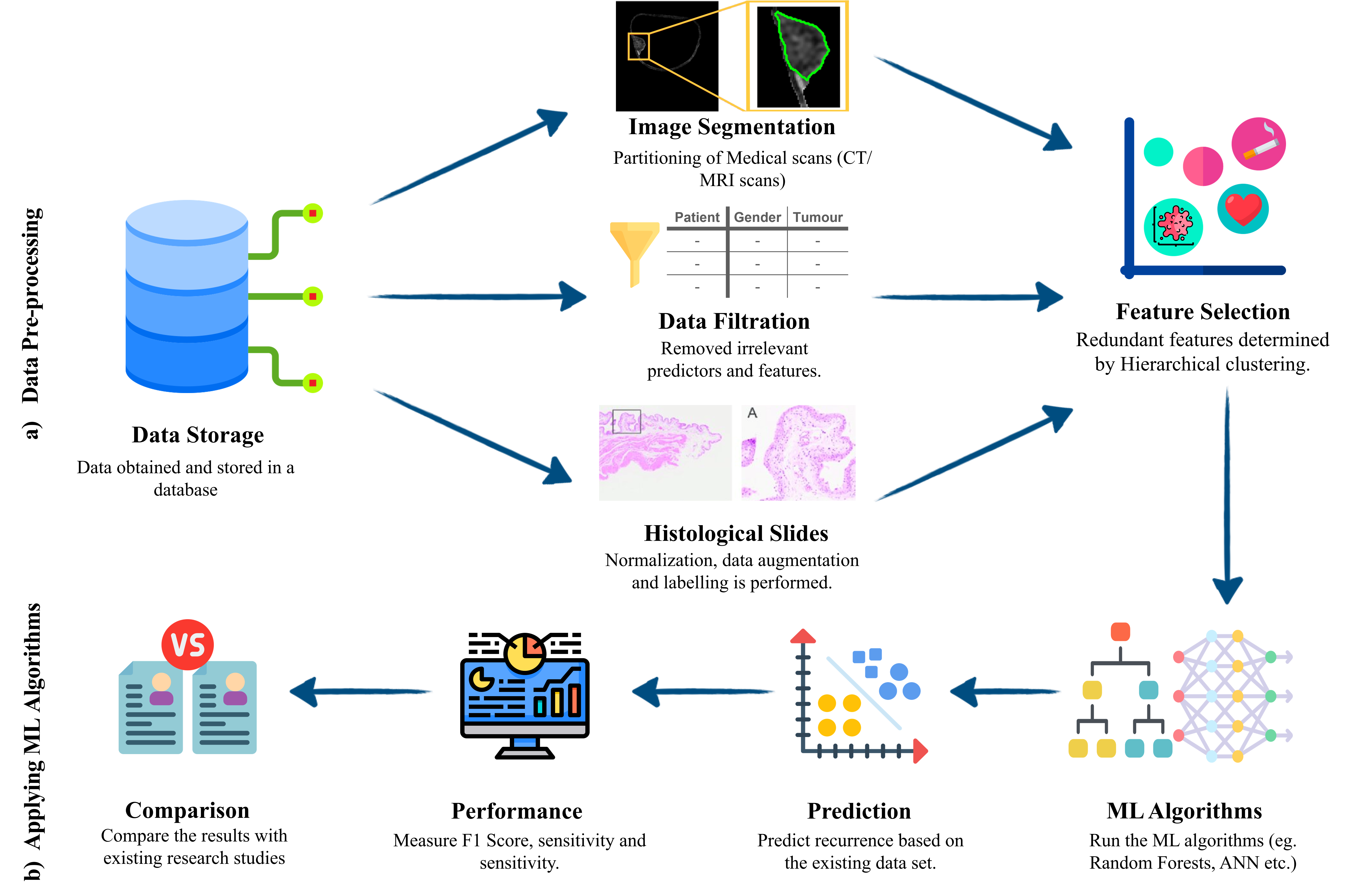}
	\caption{Machine learning workflow model for bladder cancer prediction. (a) pre-processing steps –Data stored in a secure database goes through image segmentation or filtration, depending on the data type. Then feature selection is applied to identify useful features in the data while discarding the redundant and unimportant features.  (b) application of  ML algorithms -  ML algorithms are selected and trained, prediction is made, and performance is evaluated using the most suitable metrics. Then a robust comparison is made to deduce added benefit and the superiority over the existing frameworks.}
	\label{BC_workflow}
\end{figure*}

Figure~\ref{BC_workflow} depicts a typical bladder cancer prediction workflow using an ML model, consisting of data pre-processing and ML algorithm application. The process begins with secure data storage, followed by data pre-processing, where irrelevant or incomplete data is removed and medical scans are segmented to focus on key areas like tumour lesions. Redundant features are identified and eliminated, often using clustering techniques. Then, ML algorithms are trained with this processed data for cancer recurrence prediction. The model's performance, evaluated by accuracy, precision, recall, and F1-score, may lead to retraining for improved standards. {While accuracy measures the overall correctness of the model, precision indicates the proportion of positive cases correctly identified, avoiding false positives. Similarly, recall reflects how many actual positives were correctly identified, avoiding false negatives. The F1-score combines precision and recall into a single metric, balancing both false positives and false negatives}. Finally, the model's effectiveness is compared with existing methods to assess any advancements in bladder cancer recurrence prediction and clinical detection benefits.

Accurately assessing the risk and predicting the occurrence and recurrence of NMIBC early is crucial for effective treatment and management of NMIBC. ML-based diagnosis and predictive systems have been proven valuable in tumour detection, bladder segmentation, and NMIBC identification \cite{shkolyarAugmentedBladderTumor2019,chaBladderCancerSegmentation2016}. Moreover, AI approaches also have been applied to tumour staging, grading, survival rate prediction, response to chemotherapy, and recurrence rates, all of which are essential in personalised NMIBC management \cite{jansenAutomatedDetectionGrading2020,wangRadiomicsAnalysisMultiparametric2019,mucakiPredictingResponsesPlatin2019a}. 

However, as of this date, no reliable AI algorithm is capable of accurately predicting NMIBC recurrence and improving management through a combination of the aforementioned markers. By reviewing these studies, we aim to pave the way for researchers to develop highly precise AI-based NMIBC recurrence prediction systems, facilitating optimal personalised management for early-stage NMIBC.  

\section{Methods}
\subsection{Search Strategy}
\begin{figure*}
	\centering
	\includegraphics[width=1\textwidth]{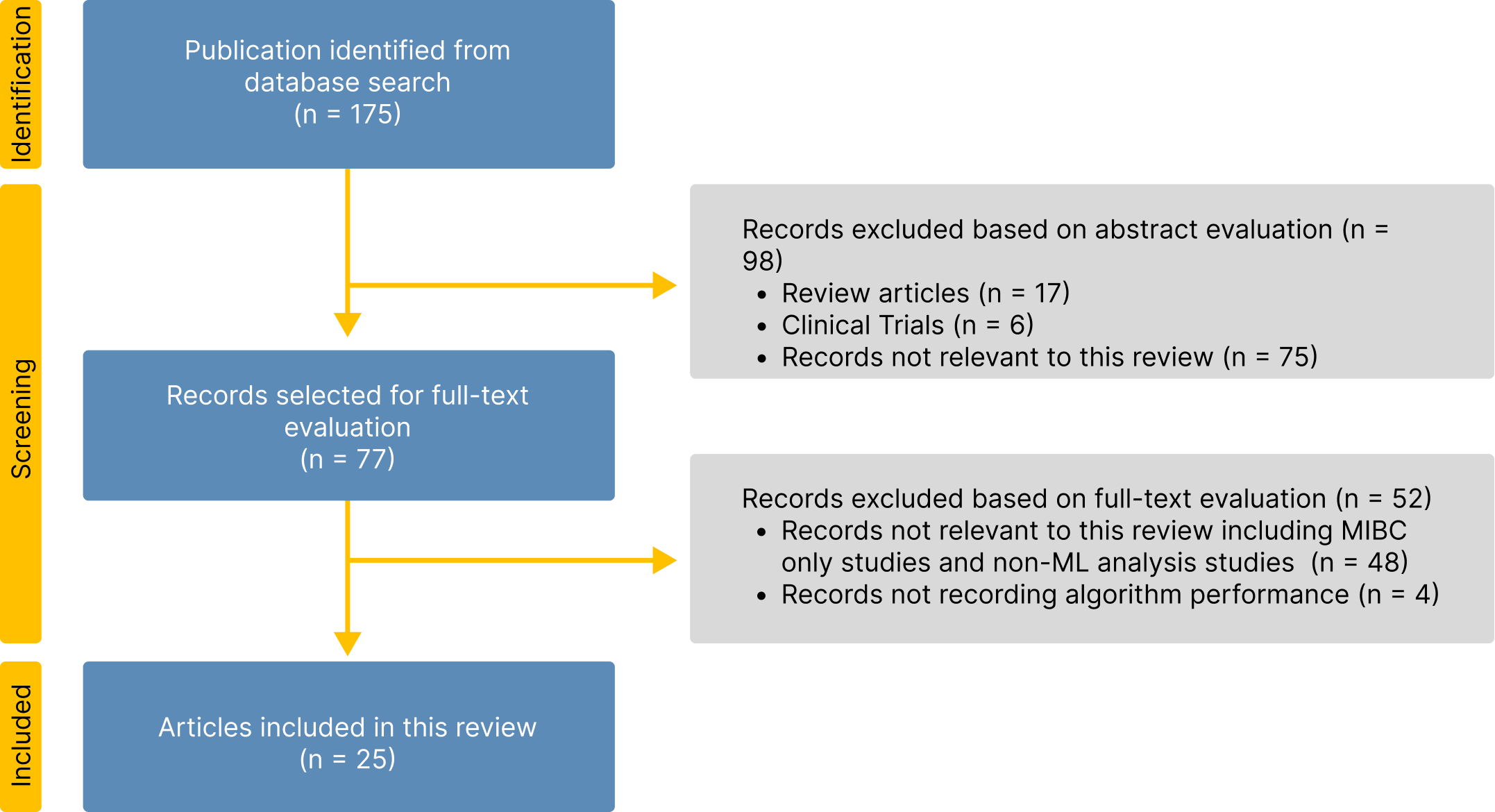}
	\caption{PRISMA flow diagram showing search methodology, inclusion and exclusion criteria.}
	\label{flow_diagram}
\end{figure*}

In this systematic review, we survey studies published up to October 2024, focusing on highly-ranked articles related to the implementation of ML and AI in bladder cancer prediction. We searched databases such as PubMed, IEEE Xplore, ScienceDirect (Elsevier), Springer, Nature, and MDPI. We used a combination of keywords that matched the scope of the survey such as "Bladder Cancer OR Non-Muscle Invasive Bladder Cancer OR NMIBC" AND "Artificial Intelligence" OR "Machine Learning" OR "Neural Networks" AND "Prediction" OR "Predictor" AND "Recurrence". Our search yielded a total of 175 {unique studies }. We reviewed the 175 articles on ML for NMIBC, focusing on those used for recurrence prediction. Upon abstract evaluation, 98 articles were discarded according to the exclusion criteria defined below. A further 46 studies were excluded based on full-text analysis. A total of 25 studies were finally selected for in-depth analysis within this review. Our study adhered to the guidelines set forth by the Preferred Reporting Items for Systematic Reviews and Meta-Analyses (PRISMA) guidelines \cite{liberatiPRISMAStatementReporting2009}. The PRISMA flowchart is depicted in Figure~\ref{flow_diagram}.

\subsection{Inclusion Criteria}
The inclusion criteria for this systematic review were defined to ensure the relevance and quality of the studies selected for analysis. The criteria are as follows: (1) original research articles, including peer-reviewed journal papers and conference proceedings; (2) studies specifically focused on the application of ML techniques for the prediction of NMIBC recurrence; (3) research exclusively addressing NMIBC recurrence or survival outcomes; (4) publications available in the English language. (5) Articles that report specific performance metrics (e.g., accuracy, sensitivity, specificity, AUC) to evaluate the effectiveness of the proposed ML models.

\subsection{Exclusion Criteria}
The exclusion criteria for this systematic review were designed to eliminate studies outside the scope of NMIBC recurrence prediction. The following were excluded: (1) studies focused solely on diagnosis without addressing recurrence or survival prediction; (2) non-original research, including systematic reviews, meta-analyses, commentaries, editorials, and case reports; (3) articles lacking full-text access or sufficient methodological details; (4) studies using only traditional statistical methods without machine learning; (5) research on cancer types other than NMIBC (such as Gall Bladder Cancer, prostate cancer, lung cancer) or studies that did not differentiate between muscle-invasive and non-muscle-invasive bladder cancer.

\section{Comparative Analysis of AI Models for NMIBC Recurrence Prediction}
\label{sec_V}

Accurate prediction of bladder cancer recurrence is critical for guiding treatment strategies and optimising patient care. Recent breakthroughs in AI have ushered a new era of possibilities, offering unprecedented opportunities to refine and improve the prediction of recurrence as well as the occurrence of various fatal diseases such as cancers. By harnessing advanced ML algorithms and integrating diverse markers such as radiomic, clinical, histopathological, and genomic data, AI-based approaches have demonstrated remarkable potential in unravelling the intricate nature of bladder cancer recurrence. In this section, we explore the pivotal studies and advancements that showcase the transformative role of AI in bladder cancer recurrence prediction, shedding light on its promises, challenges, and future prospects. 

To provide a clear understanding and a structured walkthrough of the studies, this section is categorised into four subsections based on similar methodologies as follows:

\begin{itemize}
	\item Models using Imaging and Morphological Features
	\item Models using Genomic and Protein Markers
	\item Models using Clinical, Treatment and External Factors
	\item Models using a Combination of Feature Types listed above
	
\end{itemize}

To provide a consolidated view of research in this interdisciplinary field, we have provided the Table~\ref{tab:recurrence_studies_full}  which offers a comprehensive overview of the various studies that have used ML techniques for NMIBC recurrence prediction. We provide the primary objectives of each study, the specific prediction tasks undertaken, the patient cohorts involved, the variables and data modalities employed, the chosen ML models, and the resultant performance metric.

\subsection{Models using Imaging and Morphological Features}
Imaging-based ML models have become central to predicting NMIBC recurrence, leveraging data from pathology slides, radiomics, and CT scans. These approaches provide detailed insights into tumour morphology and cellular patterns, often revealing features that traditional methods may overlook. Studies in this category use deep learning and other ML techniques to enhance prognostic accuracy by analyzing both tissue samples and imaging modalities.

The analysis of pathology slides has been a cornerstone of many ML models for NMIBC recurrence prediction, offering insights into tissue morphology and cellular characteristics. Chen et al. \cite{chenClinicalUseMachine2021} used machine learning on Hematoxylin and Eosin-stained images from 514 patients, a substantial sample size for bladder cancer research. Unlike other approaches that use nuclear extraction techniques, they applied LASSO with 10-fold cross-validation to identify 22 bladder cancer-related and 18 survival-related image features. Their diagnostic model achieved strong performance, with Area Under the Receiver Operating Curve (AUROC) values of 94.1\%, effectively distinguishing bladder cancer from normal tissues and glandular cystitis. The ML-based risk score served as an independent predictor for survival, enhancing prediction accuracy for 1-, 3-, and 5-year overall survival by more than 10\%. However, the use of non-uniform median cut-off values for high-risk scores limits consistency, indicating a need for further standardization and validation in future prospective trials before broad clinical adoption.

Building on this approach of using pathology slides, Tokuyama et al. \cite{tokuyamaPredictionNonmuscleInvasive2022} explored the predictive value of nuclear atypia in NMIBC, focusing on the morphological characteristics of cancer cells derived from transurethral resection specimen. They utilised ML to predict the recurrence of NMIBC based on nuclear atypia (abnormalities in the nuclei of cancer cells) extracted from transurethral resection specimens. Using SVM and RF algorithms on a dataset of 125 patients, the authors derived quantitative morphological features from regions of interest on Hematoxylin and Eosin-stained slides. This involved the application of a nuclear extraction process using software programs "Ilastik" and "YOLO v3" for the segmentation of individual nuclei. The SVM-based model achieved a 90\% probability of predicting NMIBC recurrence within 2 years post-TURBT, while the RF-based model achieved 86.7\%. Despite these promising results, the study's limitations included a relatively small sample size, the potential for bias in the nuclear extraction process, and the limited generalizability due to the exclusion criteria.

Drachneris et al. \cite{drachnerisCD8CellDensity2023a} provided a new perspective by analyzing immune cell density gradients. They used CD8+ cell-density gradients to predict Recurrence-free survival in NMIBC patients post-BCG therapy. CD8+ are a type of white blood cell that is crucial for fighting infections or cancer. Traditional methods merely count the total number of immune cells, while this study analyses the spatial distribution (or the cell density gradients), which provides deeper insight into the cancer's immune response. By combining the HALO AI Densenet v2 classifier (a deep learning-based image analysis platform designed for digital pathology) and multivariable Cox regression models, the study creates a robust model with a C-index of 0.74. However, the paper recognises that broader, prospective validations and explorations of other immune factors are required to further validate the methods. This study shows that a fresh perspective on established methods can enhance bladder cancer prognosis accuracy. 

The use of cellular features extends to urine cytology as well. The study by Levy et al. \cite{levyExaminingLongitudinalMarkers2023} used a Deep learning ML tool called AutoParis-X to predict bladder cancer recurrence risk from 1,259 urine cytology images. The prediction model achieved good accuracy (C-index 0.77), outperforming models using standard cytological assessment alone. It was found that the model worked best when it looked at samples from the first 6 months after the initial cancer diagnosis. This model was trained on retrospective data from 159 patients from only one hospital. The small number of patients, single-centred and retrospective analysis results in concerns about the model's generalizability. 

Pathomics, which integrates patch-level and whole-slide image analyses, represents another advancement in pathology-based models. Wang et al. \cite{wangPredictionNonmuscleInvasive2024} introduced a novel pathomics model using deep learning for predicting NMIBC recurrence with high accuracy. This study aimed to address the challenge of early recurrence prediction in NMIBC by leveraging deep learning to analyze pathology images. The model, developed using a two-phase approach—patch-level prediction followed by whole slide image (WSI)-level prediction—achieved a strong performance with an AUC of 0.860 in the test cohort. Transfer learning was employed to generalize the model across different datasets, while model interpretability was improved through visualization techniques, helping clinicians understand the predictions. However, the use of pathology images, while innovative, requires digital pathology infrastructure that may not be available in all clinical settings. The model's excellent performance suggests it could be a valuable tool in clinical practice, but external validation in larger and more diverse patient populations is needed.

Shifting from pathology images to radiomics, Xu et al. \cite{xuPredictiveNomogramIndividualized2019} shifted the focus towards integrating radiomics features from MRI scans, offering a novel approach to personalized risk assessment. Xu et al. conducted a study which aimed to devise a personalised tool for estimating the two-year recurrence risk of bladder cancer. A model incorporating both radiomics features extracted from MRI scans and clinical factors was built using data from 71 patients. The model utilised ML methods like SVM-based recursive feature elimination. This nomogram, relying on muscle-invasive status and an radiomics-derived score, showed promising accuracy (80.95\% in validation) and a high AUROC value of 0.838. Despite these strong results, potential bias due to retrospective design and single-centre sampling remain as the limitations. Future investigations could explore the potential of additional factors, currently omitted due to incomplete data, in enhancing the predictive power of the model. Moreover, the role of different radiomics features in predicting lymph node status, crucial for prognosis, was recommended for future research. 

Building on this, Huang et al. \cite{huangMultiparametricMRIBased2024} demonstrated the potential of combining multiparametric MRI and Deep learning for predicting bladder cancer recurrence. In their study, the authors developed a clinical-radiomics deep learning model that integrated radiomics features from multiparametric MRI with deep learning and clinical data to predict the 5-year recurrence risk in NMIBC patients. With a dataset of 191 patients, the model outperformed traditional clinical models, achieving an AUC of 0.909. The use of SHapley Additive ExPlanations further enhanced the interpretability of the model, showing that radiomics features contributed significantly to prediction accuracy. However, the approach relies heavily on expensive and advanced MRI technology, which may not be accessible in many clinical settings. Furthermore, while SHAP values increase interpretability, the inherent complexity of radiomics and the deep Learning models still creates a “black box” issue, limiting clinician trust in the model's outputs.

CT imaging has also been explored for recurrence prediction. Wang et al. \cite{wangDeepLearningSignature2023} developed a deep learning model using multiphase enhanced CT images to predict bladder cancer recurrence, showing substantial clinical promise. This multi-center study involved 874 patients from four centers and used CNNs to develop a signature capable of predicting recurrence risk. The model demonstrated excellent performance, with an AUC of 0.889 and a concordance index of 0.869. It outperformed traditional clinical models and staging systems, indicating its potential utility in guiding personalized treatment strategies for NMIBC patients. Despite the robust performance of the deep learning model, it relies on high-resolution imaging data, which may limit accessibility in less technologically advanced medical centers. Further prospective validation and exploration of how the deep learning model integrates with clinical decision-making would be beneficial.

Finally, specialized clinical features, such as intravesical prostatic protrusion (IPP), have been studied as independent predictors of recurrence. Lee et al. \cite{leeIntravesicalProstaticProtrusion2021} investigated the role of IPP in predicting NMIBC recurrence {using imaging data obtained from preoperative CT urography}. By analyzing the severity of IPP in 122 male NMIBC patients, the study demonstrated that structural bladder and prostate features significantly influence recurrence risk. Severe IPP ( $\geq 5 mm $) was observed in 27\% of patients and was associated with a 2.6-fold increased risk of recurrence. Using SVM, incorporating IPP improved NMIBC recurrence prediction by 6\%. Kaplan-Meier analysis showed that severe IPP negatively impacted recurrence-free especially in high-risk patients. However, limitations such as the small sample size, retrospective design, and lack of post-void residual data suggest further research is needed to explore the relationship between IPP, ageing, and bladder conditions. These findings position IPP as an independent risk factor for recurrence and its potential role in prognosis.

\subsection{Models using Genomic and Protein Markers}

A multitude of studies have attempted to predict NMIBC recurrence, each employing unique yet complementary methodologies using biological insights combined with ML. The journey begins with Zhao et al.'s \cite{zhaoPredictionPrognosisRecurrence2022} paper, which focuses on ECM-related genes' prognostic potential in Bladder Cancer. By using expression data from multiple datasets, {six ECM-related genes were identified }: CTHRC1, MMP11, COL10A1, FSTL1, SULF1, COL5A3. An AUC of 0.76 was achieved in recurrence prediction by training many ML models, including Generalised Linear Models, K-Nearest Neighbours, SVMs and Random Forests, on 675 non-recurrent and 285 recurrent bladder cancer patients. The study demonstrates the potential of ML and ECM gene signatures for recurrence prediction in bladder cancer, though larger sample sizes would help validate the approach. Limitations include few normal samples for comparison and lack of extensive validation of individual gene trends.

Another genomic study, this time by Cai et al. \cite{caiPrognosticRoleLoss2010}, analysed loss of heterozygosity (LOH) on chromosome-18 in 65 patients with NMIBC and 43 controls. Loss of heterozygosity occurs when one of the two copies (alleles) of a gene, inherited from each parent, is lost or inactivated, which can lead to cancer if the affected gene's role is to suppress tumours. At multivariate analysis, LOH on Chr 18 (P=0.002) and the number of lesions (P=0.03) were identified as independent predictors of recurrence-free probability. ANNs were used to confirm the multivariate analysis but the performance metrics were not mentioned. This paper, published in 2010, did not gain widespread adoption due to factors such as small study sample size; robustness of other recurrence predictors (FGR3 \cite{sikicPrognosticValueFGFR32021a}, Ki-67 \cite{bertzCombinationCK20Ki672014} and NMP22 \cite{wangEvaluationNMP22BladderChek2017,ponskySCREENINGMONITORINGBLADDER2001}).

Urbanowicz \cite{urbanowiczRoleGeneticHeterogeneity2013a} also aimed to uncover patterns of genetic associations with bladder cancer by applying an ML classifier system called {AF-UCS (Attribute Feedback-sUpervised Classifier System) }. { The algorithm aimed to validate findings that specific SNPs in DNA repair genes, such as XPD (Xeroderma pigmentosum group D) codon 751 and 312, along with SNPs in other DNA repair genes, are predictive of bladder cancer risk when considered alongside smoking pack-years.} Their model had an accuracy of 0.66, indicating a low to moderate level of predictive performance. Since this publication in 2013, there has been limited follow-up potentially due to other ML methods having superior accuracy, and the other models not requiring specialised expertise. Additionally, large-scale Genome-Wide Association Studies are now more common than focused candidate gene approaches \cite{wuGenomewideAssociationStudies2009, menonFindingPlaceCandidate2021, garcia-closasGenomewideAssociationStudy2011, rafnarGenomewideAssociationStudy2014, wangGenomeWideAssociationStudy2016}.

Building on the growing interest in genomic predictors, Maturana et al.  \cite{lopezdematuranaPredictionNonmuscleInvasive2016} published a study three years later that focused on predicting outcomes of NMIBC patients using genomic SNP profiles in conjunction with clinico-pathological prognosticators. The study utilized Bayesian learning methods, including sequential threshold models and LASSO, and evaluated 822 NMIBC patients followed up for over a decade. The genomic models yielded AUROC values ranging from 0.55 to 0.62, while clinico-pathological models performed slightly better, with AUROC values between 0.57 and 0.76. These results indicate that SNP profiles alone are poor predictors of NMIBC recurrence and progression, and their inclusion in clinico-pathological models adds limited value. The limitations of the study include a relatively small sample size in certain subgroup analysis and the limited predictive ability of common SNPs in NMIBC outcomes.

Chang et al. \cite{changComprehensiveUrinaryProteome2024} utilized urinary proteomic profiling to develop a non-invasive method for diagnosing and monitoring bladder cancer recurrence. By analyzing urine samples from 279 patients, the authors used a multi-support vector machine-recursive feature elimination (mSVM-RFE) algorithm to identify 13 protein markers for diagnosis and 11 markers for recurrence monitoring. The diagnostic model achieved high sensitivity (90.9\%) and specificity (73.3\%), while the recurrence monitoring model reached 75\% sensitivity and 81.8\% specificity. This study's strength lies in its non-invasive nature, providing a practical alternative to invasive procedures like cystoscopy. However, the reliance on high-resolution mass spectrometry and the single-center design may limit the generalizability of the findings. The model’s utility is promising, but larger, multi-center validation studies are required to ensure its efficacy in diverse clinical settings.

Frantzi et al.'s study \cite{frantziDevelopmentValidationUrinebased2016a} formulated and validated urine-based biomarker panels for primary and recurrent bladder cancer detection using capillary electrophoresis-mass spectrometry (CE-MS). {Their methodology combined statistical analysis and machine learning: initial statistical tests identified significant peptide biomarkers, which were further refined using SVMs via MosaCluster software to create and optimise high-dimensional biomarker panels.} Case-control comparisons across multicentre cohorts identified the biomarkers, with the primary panel achieving an AUROC of 0.87 (91\% sensitivity, 68\% specificity) and the recurrent panel attaining an AUROC of 0.75 (88\% sensitivity, 51\% specificity) during independent validation. By incorporating all available biomarkers, optimised panels improved performance further, achieving AUROCs of 0.88 for primary and 0.76 for recurrent cancer. These findings highlight the potential for non-invasive urine-based tests to complement or reduce the need for invasive cystoscopy in bladder cancer diagnosis and monitoring. Despite these promising results, the study was limited by unadjusted confounding variables, such as tumour size and hematuria, and by its cross-sectional design. Further research is needed to confirm their clinical utility and performance in real-world settings.

This study by Krochmal et al. \cite{krochmalUrinaryPeptidePanel2019} analysed CE-MS peptidomics data (detected types and amounts of peptides in a biological sample) from 98 bladder cancer patients to develop an ML model to predict recurrence. With a training set of 50 patients, Cox regression identified 36 peptides predictive of relapse which were then input into a Random Forest model. The training set produced an accuracy of 100\%, while test set accuracy is not mentioned. The limited size of the training dataset (n=48), and the unreported test accuracy, calls into question the study's generalizability. A comparison with established clinical tools would have highlighted any advancements over existing methods.

Mucaki et al. \cite{mucakiPredictingResponsesPlatin2019a} applied biochemically-inspired machine learning (ML) models, specifically supervised SVMs, to predict responses to chemotherapy agents, including cisplatin, a critical drug in bladder cancer treatment. The models were described as "biochemically-inspired" because they incorporated genes with established biological relevance to the mechanisms of action and resistance of cisplatin, ensuring the selection of features grounded in prior biochemical knowledge rather than relying purely on data-driven approaches. These gene signatures were designed to reflect pathways involved in apoptosis, DNA repair, and drug transport, enhancing interpretability and biological validity. Using a dataset of 90 cancer patients, the cisplatin-specific gene signature achieved 71.2\% accuracy in predicting bladder cancer recurrence, with exceptional performance in non-smokers (100\%) and 79\% accuracy in smokers. The study also utilised ensemble averaging across multiple thresholds to improve model robustness. However, a key limitation was the reliance on breast cancer cell line data to train the SVM models, which limits their applicability to bladder cancer-specific contexts, as no retraining was performed on bladder cancer datasets. While this work highlights the promise of AI-driven approaches for identifying recurrence risk factors, it also underscores the need for developing bladder cancer-specific models to improve clinical relevance.

Two other studies (Zhan et al. \cite{zhanExpressionSignaturesExosomal2018a} and Gogalic et al. \cite{gogalicValidationProteinPanel2017}) explored the application of protein panels in recurrence prediction. Zhan et al. utilised MALAT1, PCAT-1, and SPRY4-IT1 biomarkers and achieved an accuracy of 81.3\% and a sensitivity and specificity of 0.625 and 0.850 respectively. Additionally, tumour stage showed a statistically significant correlation as a predictor, with PCAT-1 identified as an independent predictor. Gogalic et al. combined common clinicopathological markers with ECadh, IL8, MMP9, EN2, and VEGF biomarkers. Their model, incorporating these markers, yielded an AUROC of 0.84.

\subsection{Models using Clinical, Treatment and External Factors}

Ajili et al. \cite{ajiliPrognosticValueArtificial2014} applied ANNs to predict bladder cancer recurrence after Bacillus Calmette-Guerin immunotherapy. Using a multilayer perceptron model in MATLAB, the researchers incorporated patient characteristics, tumour attributes, and treatment details, achieving high performance, with a mean square error of 0.02634. The model accurately classified 39 out of 40 cases, yielding sensitivity of 96.66\%, specificity of 100\%, and positive/negative predictive values of 100\% and 90.9\%, respectively. However, the small sample size and difficulties in determining the optimal network topology (e.g., hidden layer nodes) limited the study. These results showcase the potential of ANNs in bladder cancer prognosis but illuminate the need for larger datasets and further validation before clinical application.

Zhang and Ma \cite{zhangPredictiveValueTotal2024} investigated the predictive value of two clinical biomarkers, CA50 and total bilirubin (TBIL), for bladder cancer recurrence using ML models. The study evaluated the individual and combined predictive performance of these biomarkers in a cohort of 345 bladder cancer patients. The results demonstrated that CA50 had an AUC of 0.602 (p = 0.038) and TBIL had an AUC of 0.585 (p = 0.014), indicating that both biomarkers were moderate predictors of recurrence. When combined, the AUC increased to 0.623 (p = 0.013), showing an improvement in predictive power, though still relatively modest. While the combination of these two biomarkers offered better performance than either one alone, the overall predictive accuracy remained limited, suggesting that additional features or biomarkers may be necessary to enhance the model's ability to predict bladder cancer recurrence. The study provides valuable insights into the role of clinical biomarkers in cancer prognosis but highlights the need for further optimization to improve predictive robustness.

Schwarz et al. \cite{schwarzRelevantFeaturesRecurrence2024} examined the role of explainability in ML models for predicting bladder cancer recurrence, enhancing their clinical utility. In this study, the authors used three ML models—SVM, gradient boosting, and ANNs—and compared them to logistic regression for predicting 2-year recurrence in urothelial carcinoma patients. Gradient boosting performed best, with an F1-score of 83.89\% and AUC of 70.82\%. To address the black-box nature of these models, the authors employed permutation feature importance and feature importance ranking measure to explain the most influential features driving predictions, such as therapeutic measures. This approach enhances the transparency of ML models, potentially increasing their adoption in clinical settings.  However, while they attempt to address the "black-box" issue using feature importance measures like permutation feature importance and feature importance ranking measure, the interpretability of these models remains limited. These feature importance methods often fail to provide clinically actionable insights, as they simply show which features contribute most to the prediction without explaining why.

\subsection{Models using a Combination of Feature Types}
\label{subsec_combination_of_features}
Integrating diverse feature types—such as imaging, morphological, clinical, and genomic markers—has proven to significantly enhance the predictive performance of AI models for NMIBC recurrence. By leveraging multiple data sources, these models offer a more comprehensive understanding of tumour behavior and patient risk profiles. This section reviews key studies that have successfully combined these feature types, highlighting their methodologies, outcomes, and potential for clinical application.

One of the earliest studies in this area, conducted by Catto et al. \cite{cattoArtificialIntelligencePredicting2003a}, aimed to compare the predictive accuracies of neuro-fuzzy modelling (NFM), neural networks (NN), and traditional statistical methods for predicting bladder cancer recurrence. The study used data from 109 NMIBC patients who were treated with  TURBT. The data was used to train and test the predictive models. The study found that both NFM and NN predicted the patients' relapse with an accuracy ranging from 88\% to 95\%, which was superior to statistical methods (71-77\%). The difference was statistically significant, indicated by the p-value of less than 0.0006. The low p-value suggests that these differences are not just due to random chance. NFM appeared better than NN at predicting the timing of relapse (p $<$ 0.073). Importantly, NFM offered a transparency advantage over NN, allowing for easier clinical validation and manipulation of input variables for exploratory predictions. This early work highlighted the potential of combining different data types to boost predictive power while maintaining interpretability.

Lucas et al. \cite{lucasDeepLearningBased2022a} further advanced this approach by incorporating both histopathological image features and clinical data to predict recurrence-free survival in NMIBC patients. Their model, built using Convolutional Neural Networks (CNNs) and Bidirectional GRUs, achieved a 1-year AUC of 0.62 (n=359) and a 5-year AUC of 0.76 (n=281). Their prediction model consisted of a multi-step process that combined features extracted from the histopathological images with the clinical data to create an overall outcome. Their model's strength lies in the autonomous analysis of histopathological data, reducing human bias and possibly capturing more nuanced features. However, the study’s relatively small, single-center dataset and extended inclusion period raised concerns about generalizability and consistency in clinical practices over time. Nevertheless, the integration of image and clinical data illustrated the power of multi-modal approaches in predicting bladder cancer recurrence.

In this study \cite{jobczykDeepLearningbasedRecalibration2022}, a deep learning-based approach (DeepSurv) was used by Jobczyk et al. to recalibrate the prediction tools for the recurrence and progression of NMIBC in a cohort of 3,892 patients. The existing risk groups, EORTC and CUETO, showed moderate performance in predicting survival outcomes. The deep learning models displayed improved accuracy. In the training group of 3,570 patients, the c-indices were 0.650 for recurrence-free survival and 0.878 for progression-free survival. In the validation group of 322 patients, the c-indices stood at 0.651 for recurrence-free survival and 0.881 for progression-free survival. The models surpassed the performance of standard risk stratification tools and demonstrated no signs of overfitting. These findings highlight the potential of deep learning models in enhancing the prediction of recurrence and progression in NMIBC, offering a valuable tool for personalised patient care.

\newpage
\onecolumn
\begin{landscape}
	\tiny
	\renewcommand{\arraystretch}{1} 
	\begin{longtable}{p{1.5cm}p{4.0cm}p{2cm}p{1.2cm}p{3.5cm}p{1.5cm}p{2.5cm}p{4cm}} 
		\captionsetup{width = 680pt}
		\caption{Summary of Studies Utilising Machine Learning Techniques for NMIBC recurrence Prediction and Assessments}
		\label{tab:recurrence_studies_full} \\
		
		\hline \hline
		\textbf{Reference} & \textbf{Objective} & \textbf{Prediction Task} & \textbf{Patients} & \textbf{Variables} & \textbf{Data Modality} & \textbf{ML Model} & \textbf{Performance}   \\
		\hline
		\endhead
		\hline
		\multicolumn{8}{c}{Continued on next page...} \\
		\endfoot
		\hline
		\hline
		\multicolumn{8}{p{680pt}}{ACCI = Age-adjusted Charlson Comorbidity Index; ANN = Artificial Neural Networks; ASA = American Society of Anesthesiologists; BCa = Bladder Cancer; BPN = Back Propagation Neural network; CCI = Charlson Comorbidity Index; CHF = Congestive Heart Failure; DL = Deep Learning; DSS = Disease-Specific Survival; ELM = Extreme Learning Machine; EMR=Electrical Medical Records; GBT = Gradient Boosting Trees; HR = Hazard Ratio; IPP = Intravesical Prostate Protrusion; KNN = K-Nearest Neighbours; LR = Logistic Regression; LinR = Linear Regression; MLP = Multilayer Perceptron; NB = Naive Bayes; NBI = Narrow Band Imaging; NFM = Neuro-fuzzy modeling; NMIBC = Non-Muscle Invasive Bladder Cancer; NPV = Negative Predictive Value; OR = Odds Ratio; OS = Overall Survival; PFS = Progression-Free Survival; PPV = Postitive Predictive Value; RBFN = Radial Basis Function Network; RELM = Regularized Extreme Learning Machine; RF = Random Forest; RFS = Recurrence-Free Survival; SNP = Single Nucleotide Polymorphism; SVM = Support Vector Machine; TCGA = The Cancer Genomic Atlas; TUR = Transurethral Resection; UE = Urinary Exosome; WL = White Light.}
		
		\endlastfoot
		Tokuyama et al. \cite{tokuyamaPredictionNonmuscleInvasive2022} & Predict NMIBC recurrence & Recurrence & 125 & Nuclear atypia from TUR specimens & H\&E stained slides & SVM, Random Forests & SVM: 90.0\%, \newline RF: 86.7\% \\
		\hline 
					
		Chen et al. \cite{chenClinicalUseMachine2021} & Diagnostic and prognostic models for BCa based on H\&E images & Diagnosis and survival & 514 & H\&E stained images & Digital pathological images & LASSO-Cox hazard model & Diagnostic model: \newline AUROC: 89.2\% - 96.3\%; \newline \newline Prognostic model: \newline HR = 2.09 (TCGA cohort), \newline HR = 5.32 (General cohort) \\
		\hline 

		Xu et al. \cite{xuPredictiveNomogramIndividualized2019} & Personalised tool for BCa recurrence risk & Recurrence risk & 71 & Radiomics and clinical factors & MRI and clinical data & SVM with feature elimination & Accuracy: 81.0\% \newline AUROC: 83.8\% \\
		\hline  

		Levy et al. \cite{levyExaminingLongitudinalMarkers2023} & Predict bladder cancer recurrence using imaging features from urine cytology & Recurrence & 135 & Imaging features from urine cytology exams & Urine cytology slide images & AutoParis-X (NN)& C-index: Up to 0.77 \\
		\hline

		Huang et al. \cite{huangMultiparametricMRIBased2024} & Predict 5-year recurrence risk in NMIBC using MRI and deep learning & Recurrence & 191 & Clinical, radiomics, and deep learning features & Multiparametric MRI & Clinical and Radiomics & AUC: 0.909, C-index: 0.804 \\
		\hline

		Wang et al. \cite{wangPredictionNonmuscleInvasive2024} & Predict recurrence of NMIBC using deep learning on pathology images & Recurrence & 210 & Pathological and clinical features & Pathology slide images & Deep learning (patch-level, WSI-level) & AUC: 0.860\\
		
		\hline

		Lee et al. \cite{leeIntravesicalProstaticProtrusion2021} & Significance of IPP in NMIBC prognosis & Prognosis & 122 & IPP, age, BPH and other clinicopathological variables & Clinical data & SVM & Accuracy: 80.0\% \\
		\hline
		 
		Zhao et al. \cite{zhaoPredictionPrognosisRecurrence2022} & Predict prognosis and recurrence of bladder cancer using ECM-related genes & Prognosis and Recurrence & 960 & Six ECM-related genes, FSTL1, stage, age, gender & Genetic and clinical data & GLM, KNN, SVM, Random Forests & AUC: 76\%\\
		\hline

		Cai et al. \cite{caiPrognosticRoleLoss2010} & Evaluate prognostic role of LOH on chromosome 18 in low-risk NMIBC & Recurrence & 108& LOH on chromosome 18, number of lesions, clinico-pathological factors & Clinical data & Univariate and multivariate analyses, NN analysis & Multivariate analysis found LOH as an independent predictor of recurrence-free probability; NN performance metrics not mentioned.\\
		\hline
		
		Wang et al. \cite{wangDeepLearningSignature2023} & Predict bladder cancer recurrence using multiphase CT images & Recurrence & 874 & Clinical and imaging features & Multiphase enhanced CT images & Convolutional Neural Networks & AUC: 0.889, C-index: 0.869 \\
		\hline

		Urbanowics et al. \cite{urbanowiczRoleGeneticHeterogeneity2013a} & Identify genetic and environmental factors in bladder cancer susceptibility & Risk and Survival time & 914& DNA repair gene SNPs, smoking history& Genetic and clinical data& AF-UCS (Learning Classifier System)& Accuracy: 66\%\\
		\hline
		
		Maturana et al. \cite{lopezdematuranaPredictionNonmuscleInvasive2016} & SNP impact on NMIBC prognosticators &Prognosis & 995 & 171,304 SNP, 6 clinical-pathological indicators & Clinical and genetic data & Bayesian learning with LASSO & Time to First Recurrence AUC: \newline Clinico-pathological = 0.62 \newline SNP-only = 0.55 \newline Combined = 0.61. \newline Time to progression AUC: \newline Clinico-pathological = 0.76 \newline SNP-only = 0.58 \newline Combined = 0.76\\
		\hline

		Krochmal et al. \cite{krochmalUrinaryPeptidePanel2019} &Predict bladder cancer recurrence using Urinary peptide panel&Recurrence & 98&Urinary peptide panel &Urine samples & Random Forests & AUC: 0.828 \\
		\hline

		Frantzi et al. \cite{frantziDevelopmentValidationUrinebased2016a} & Develop and validate urine-based biomarker panels for primary and recurrent BCa detection & Biomarker panels for detection & 481 & Peptide biomarkers & Urine samples &  & Primary panel:\newline Accuracy 76\%, AUC 0.77 \\
		\hline

		Zhan et al. \cite{zhanExpressionSignaturesExosomal2018a} & Develop a urinary exosome derived lncRNA panel for BCa diagnosis and recurrence & Diagnosis and Recurrence & 368 & MALAT1, PCAT-1, SPRY4-IT1 lncRNA expressions & LncRNA data & Multivariate Logistic Regression & AUC: 0.813 (validation) \\
		\hline 

		Gogalic et al. \cite{gogalicValidationProteinPanel2017} & Validate a protein panel for noninvasive detection of recurrent NMIBC & Diagnosis and recurrence & 45 & Biomarkers (ECadh, IL8, MMP9, EN2, VEGF, past recurrences, BCG therapies, stage at diagnosis) & Urine samples & Logistic regression with LASSO & Highest AUC: 0.96\\
		\hline
		
		\hline
		Zhang and Ma \cite{zhangPredictiveValueTotal2024} & Identify biomarkers for bladder cancer recurrence using ML & Recurrence & 345 & TBIL, CA50, and 34 clinical parameters & Clinical and biomarker data & Decision tree, Random Forest, AdaBoost, GBM, XGBoost & Best model AUC: 0.623 \\
		\hline
		
		Chang et al. \cite{changComprehensiveUrinaryProteome2024} & Develop non-invasive urine-based proteomic biomarkers for recurrence & Recurrence & 279 & 11 Urinary proteins & Urine proteomic data & mSVM-RFE & AUC: 0.784 \\
		\hline 
		
		Mucaki et al. \cite{mucakiPredictingResponsesPlatin2019a} & Predicting chemotherapy responses & Recurrence and remission & 54 & Gene expression, clinical metadata & RNA-seq, microarray & SVM & Recurrence Acc: 71.0\% (cisplatin), 60.2\% (carboplatin), 54.5\% (oxaliplatin) \newline Remission Acc: 59\%, 61\%, and 72\% \\
		\hline 

		Drachneris et al. \cite{drachnerisCD8CellDensity2023a} & Predict recurrence-free survival in NMIBC post BCG therapy & Recurrence-free survival & 157 &CD8+ cell density indicators (Immunodrop, Center of Mass), tumour stage, tumour grade& Histopatho-logical slides&Multivariable Cox regression&C-index: 0.7837\\
		\hline

		Ajili et al. \cite{ajiliPrognosticValueArtificial2014} & Predict BCa recurrence post-BCG immunotherapy & Recurrence classification & 308 & Clinical variables: Age, gender, tumour stage, grade, size, multiplicity, smoking, CD34 expression & Histopatho-logical data & Neural Networks & Sensitivity: 96.66\%, Specificity: 100\%, PPV: 100\%, NPV: 90.9\% \\
		\hline 

		Catto et al. \cite{cattoArtificialIntelligencePredicting2003a} & To compare the predictive accuracies of NFM, NN, and traditional statistical methods for the behaviour of BCa & Recurrence (Occurrence and Timing) & 109 & Experimental molecular biomarkers (p53, mismatch repair proteins) and conventional clinicopathological data & Clinical and molecular data & NFM, Neural Networks, Logistic Regression and Linear Regression & NFM and NN Accuracy: 88-95\%\newline LR and LinR accuracy = 71-77\% \\
		\hline

		Lucas et al. \cite{lucasDeepLearningBased2022a} & Predict NMIBC recurrence & 1- and 5-year RFS & 359 & Digital histopathology slide data and clinical data & Digital histopathology slides and clinical records & Convolutional Neural Networks & AUC: 0.62 (1-year) and 0.76 (5-year) \\
		\hline

		Jobczyk et al. \cite{jobczykDeepLearningbasedRecalibration2022} & Recalibrate prediction tools for the recurrence and progression of NMIBC & RFS and PFS prediction & 3,892 & Gender, age, T stage, histopathological grading, tumour burden and diameter, EORTC and CUETO scores, intravesical treatment & Clinical data & DeepSurv (Deep Neural Networks) & RFS: C-index 0.65, \newline PFS: C-index 0.88 \\
		\hline 


		Schwarz et al. \cite{schwarzRelevantFeaturesRecurrence2024} & Improve explainability in ML for predicting UBC recurrence & Recurrence & 1,944 & Clinical and therapeutic features & Tabular clinical data & SVM, Gradient Boosting, ANN & AUC: 70.82\%, F1-Score: 83.89\\

	\end{longtable}
\end{landscape}

\section{Discussion}
 
Our comprehensive review of 25 ML-based studies highlights a growing trend in the adoption of ML-based approaches for NMIBC prediction and management. This demonstrates the potential of these approaches to drive a transformative shift in bladder cancer care. Our work offers a distinct contribution as we move beyond surface-level summaries to conduct a rigorous, in-depth evaluation of each study’s methodologies. We prioritize a detailed interpretation of the reported model performances, assess the robustness of sample sizes, explore the diversity of data modalities, and critically analyze the specific prediction tasks—key aspects that are often overlooked in other reviews. By focusing on these crucial factors, we uniquely uncover strengths and weaknesses that were previously missed, providing a deeper, more nuanced understanding of each study's true contribution to the field.
 
\begin{figure}
	\centering
	\includegraphics[width=0.6\textwidth]{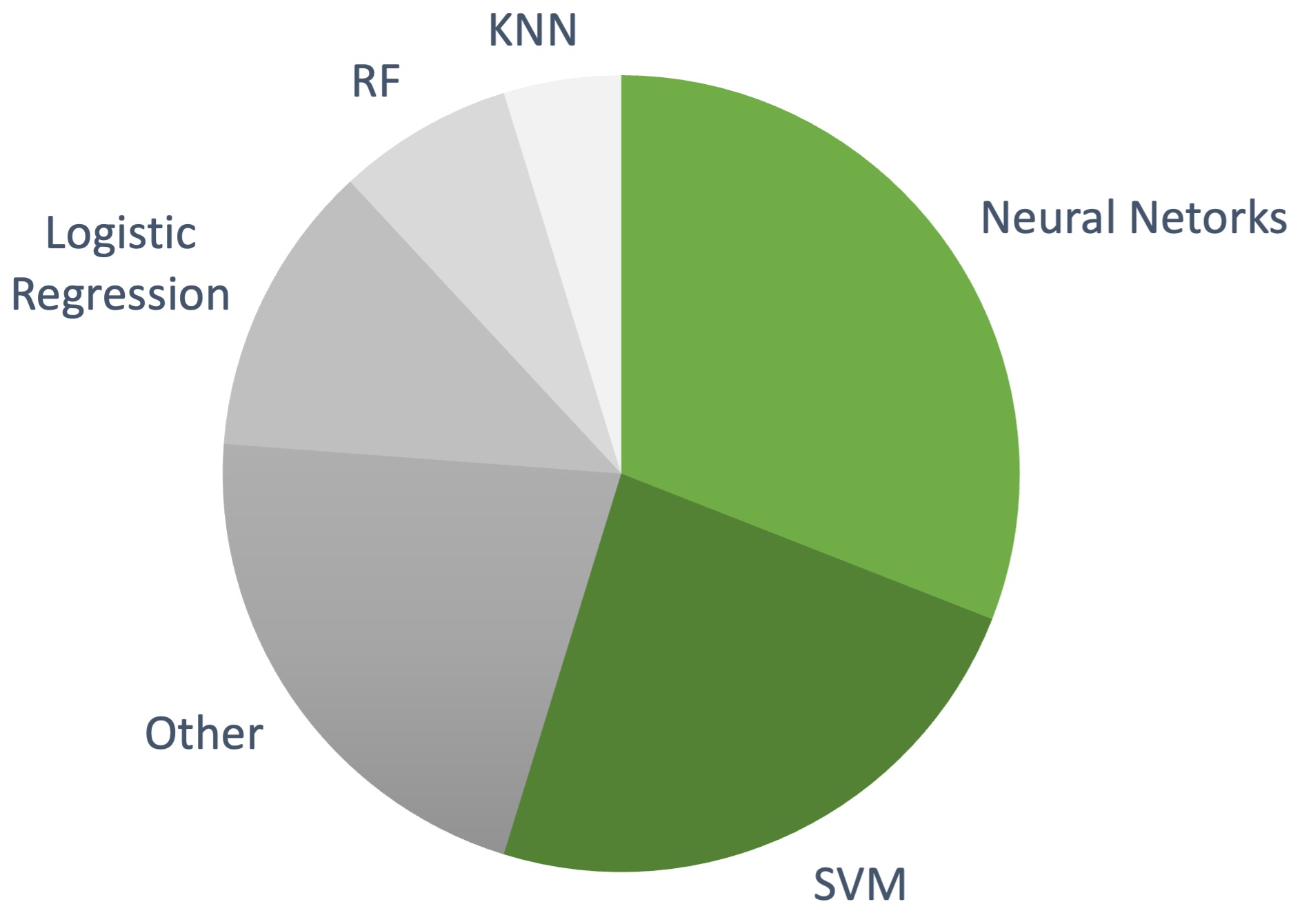}
	\caption{The distribution of machine learning models used in NMIBC recurrence prediction studies included in this review. More complex models, such as NNs and SVMs, dominate the research landscape, while simpler, interpretable models like logistic regression are used less frequently}
	\label{ML_Types}

\end{figure}

Our analysis reveals a significant evolution in the field. Early studies were confined to single data types, but as the field progressed, more advanced ML models integrating multiple modalities—such as radiomics from MRI and CT scans, genetic markers, and histopathology—showed considerable promise. This integration has led to superior predictive performance, emphasizing the importance of a holistic approach to NMIBC recurrence prediction. A significant number of studies on NMIBC recurrence prediction employ these complex ML models, as evidenced by Figure~\ref{ML_Types}. 

Among these, NNs have emerged as the most commonly used approach, featured in 10 studies \cite{levyExaminingLongitudinalMarkers2023,ajiliPrognosticValueArtificial2014,drachnerisCD8CellDensity2023a,cattoArtificialIntelligencePredicting2003a,caiPrognosticRoleLoss2010,lucasDeepLearningBased2022a, jobczykDeepLearningbasedRecalibration2022, huangBodyCompositionPredictor2023, wangPredictionNonmuscleInvasive2024, wangDeepLearningSignature2023}. These models, which incorporate clinical, pathological, and genomic markers, achieved accuracies ranging from 0.65 to 0.975. NNs are particularly suited to handling complex, multi-modal datasets and often outperform simpler models in predicting recurrence, making them an attractive option for integrating various data types. SVMs were the second most used algorithm, appearing in 9 studies \cite{xuPredictiveNomogramIndividualized2019, leeIntravesicalProstaticProtrusion2021, jobczykDeepLearningbasedRecalibration2022,tokuyamaPredictionNonmuscleInvasive2022,zhaoPredictionPrognosisRecurrence2022, mucakiPredictingResponsesPlatin2019a, krochmalUrinaryPeptidePanel2019, schwarzRelevantFeaturesRecurrence2024, changComprehensiveUrinaryProteome2024}. These studies reported overall accuracies around 0.75. SVMs are known for handling high-dimensional data well, maintaining competitive performance while offering a balance of interpretability and complexity.

By comparison, simpler models like logistic regression have shown lower performance. For example, Hasnain et al. \cite{hasnainMachineLearningModels2019} used logistic regression alongside more complex models like SVM and Random Forest, finding it lagged in predictive power. While easier to interpret, simpler models often struggle with multi-modal and high-dimensional data compared to more advanced algorithms. Another notable trend is the integration of traditional statistical models with ML techniques, such as in Jobczyk et al. \cite{jobczykDeepLearningbasedRecalibration2022}, where a Cox proportional hazards model was combined with a deep neural network. This hybrid approach allows models to retain the interpretability of traditional methods while leveraging the predictive power of modern AI techniques, a promising direction for clinical adoption.

Building on our review, we highlight the transformative potential of integrating ML-based models into NMIBC management. Below, we outline key opportunities and challenges that must be addressed for successful clinical adoption.

\subsection{Opportunities and Transformative Potential}

\begin{itemize}
		
	\item \textbf{Enhanced Accuracy and Precision:} One of the most compelling opportunities presented by ML in NMIBC management is the potential for improved diagnostic accuracy and precision. ML models can identify subtle patterns and complex relationships that may be overlooked by human observers, reducing interobserver variability and mitigating the influence of individual clinician biases. This enhanced consistency can lead to better patient outcomes through more accurate predictions of recurrence and progression.

	\item \textbf{Automation Leading to Productivity and Cost Reduction:} ML algorithms can automate complex and time-consuming tasks such as image analysis, pattern recognition, and data integration. Automation increases productivity by enabling faster processing of large volumes of data and reduces healthcare costs through more efficient resource utilization. This allows clinicians to focus more on patient care rather than administrative or repetitive tasks.

	\item \textbf{Reduction of Interobserver Variability and Clinician Bias:} Diagnoses often rely on each clinician's personal experience and are subject to interobserver variability and bias. ML models can provide consistent and objective analyses, reducing variability between clinicians and minimizing the impact of subjective judgment.

	\item \textbf{Personalized Medicine and Transferable Knowledge:} ML models facilitate personalized treatment strategies by analyzing individual patient data to predict responses to specific therapies. Tailoring treatment plans to each patient's unique profile can improve efficacy and reduce unnecessary interventions. Additionally, the methodologies developed through ML in NMIBC are transferable to muscle-invasive bladder cancer (MIBC) and other malignancies, broadening the impact of this technology. ML tools can also serve as educational resources for training new professionals, enhancing learning through explainable models that illustrate decision-making processes.
\end{itemize}

\subsection{Challenges and Considerations}
Despite the encouraging results presented in multiple studies, several significant hurdles exist before ML-based models can be successfully implemented in clinical settings. Below, we discuss key challenges identified in our review.

\begin{itemize}

	\item \textbf{Limited Generalizability and Overfitting:} AI models often struggle with the heterogeneity of NMIBC tumours in stage, grade, and molecular subtypes. Small sample sizes in many studies lead to overfitted models that are not representative of the broader NMIBC population. High predictive accuracy reported by studies like Ajili et al. is limited by factors such as retrospective designs and single-institution datasets, which affect generalizability. To address this, multi-institutional data, diverse imaging techniques, and a wide range of NMIBC presentations should be included in training datasets.
	
	\item \textbf{Complexity of Implementation in Clinical Settings:} Implementing complex multimodal models in clinical practice requires specialized expertise, which may not always be available. Multimodal models need more complex data preprocessing and tuning, making scaling difficult across centers. Additionally, the lack of transparency in some models, particularly deep learning, hinders their adoption in clinical practice.

	\item \textbf{Lack of Interpretability and the "Black-Box" Nature of AI Models:} Clinicians need models that not only provide accurate predictions but also explain their reasoning to support confident, informed medical decisions. This challenge highlights a broader issue in the field of ML in healthcare: a significant disconnect between ML research and clinical requirements. While researchers often prioritize high accuracy, favoring complex models that enhance publication appeal, clinicians prioritize models that are interpretable, even if they sacrifice \textit{some} accuracy. Bridging this gap requires a focus on developing models that balance performance with interpretability, better aligning ML innovations with the practical needs of real-world clinical applications.

	\item \textbf{Regulatory Challenges and Algorithmic Bias:} Predicting NMIBC recurrence faces challenges such as data access regulations, patient confidentiality, and ethical approvals, which can delay research and limit data availability. Additionally, ML algorithms may perpetuate biases, reducing prediction reliability for underrepresented groups. Addressing these issues requires implementing \textit{federated learning} to train models across institutions without sharing raw data, using \textit{differential privacy} to anonymize sensitive information, and ensuring NMIBC datasets reflect diverse patient demographics to improve accuracy and fairness.

\end{itemize}

\subsection{Future Directions}

To fully leverage ML in NMIBC management, future efforts should prioritize developing multimodal models, improving interpretability, integrating ML into clinical workflows, and addressing ethical and regulatory barriers.

\begin{itemize}
    \item \textbf{Multimodal Models:} Combining genomics, radiomics, and clinical data can enhance predictive power and generalizability. International collaboration is essential to assemble large, diverse datasets that capture variability in patient demographics and practices. Federated learning can enable training on multi-center data while preserving patient privacy, thus boosting model robustness.

    \item \textbf{Improving Interpretability:} To build clinician trust, models should incorporate explainable AI techniques, such as SHAP values or attention mechanisms, making predictions transparent. User-friendly interfaces with visualization tools can further support clinician interaction with model outputs, enhancing usability in practice.

    \item \textbf{Integrating ML into Clinical Workflows:} Embedding ML tools within Electronic Health Records can offer real-time decision support. Adaptive decision-support systems that refine recommendations based on clinician feedback and evolving patient data will enhance decision-making by incorporating both clinical expertise and algorithmic insights.

    \item \textbf{Regulatory Sandbox for ML Testing:} Establishing a regulatory sandbox can accelerate validation by allowing controlled testing in clinical settings with flexible oversight. This framework supports iterative model adjustments and real-time data collection, speeding up the approval process. Engaging regulatory bodies to develop NMIBC-specific guidelines will facilitate smoother pathways to clinical adoption.
\end{itemize}

\section{Conclusion}

NMIBC, with its high recurrence rate of 70\%-80\% and substantial treatment costs, demands innovative predictive solutions. Our systematic review of 25 ML-based studies highlights the transformative potential of ML in NMIBC management. We explore various ML-based frameworks that utilize radiomics, histopathological markers, clinical data, genomics, and their combinations to predict NMIBC recurrence. Studies that integrated multiple data sources demonstrated remarkable accuracy, with NNs leading the charge. Our review encompasses the usage of ML-based models, acknowledges the potential for failures, and emphasizes the need for further intensive investigations to ensure their beneficial application. Unlike previous reviews that leave researchers struggling with complex technical details, our in-depth and nuanced evaluation of each study’s methodologies offers simplified, valuable insights into the intricacies of these ML algorithms, their clinical relevance, and practical applications. The future of NMIBC management is poised for innovation, as ML models have the potential to reduce both bias and interobserver variability. However, a significant hurdle remains - the scarcity of high-quality data. Developing robust and expansive datasets through collaboration is crucial for training models that can deliver real-world impact. While ML has shown immense potential in reshaping personalized medicine, its role in predicting NMIBC recurrence is not yet the gold standard. 

\section*{Conflict of Interest Statement}

The authors declare that the research was conducted in the absence of any commercial or financial relationships that could be construed as a potential conflict of interest.

\section*{Data availability statement}
The original contributions presented in the study are included in the article. Further inquiries can be directed to the corresponding author. 

\section*{Author Contributions}
SA: Data Curation, Formal Analysis, Investigation, Methodology, Resources, Validation, Visualization, Writing – original draft, Writing – review \& editing. NS: Supervision, Writing – review \& editing. RS: Funding acquisition, Supervision, Validation, Writing – review \& editing. RH: Supervision, Validation, Writing – review \& editing. KA: Conceptualization, Funding acquisition, Investigation, Methodology, Project administration, Resources, Supervision, Validation, Writing – original draft, Writing – review \& edit.

\section*{Funding}
The author(s) declare financial support was received for the research, authorship, and/or publication of this article. This research received funding from EPSRC EP/X036006/1 to cover the publication costs.

\bibliographystyle{Frontiers-Vancouver} 

{
\small
\bibliography{review_paper_bibliography_frontier}

\begin{thebibliography}{93}
\expandafter\ifx\csname natexlab\endcsname\relax\def\natexlab#1{#1}\fi
\expandafter\ifx\csname urlstyle\endcsname\relax
  \expandafter\ifx\csname doi\endcsname\relax
  \def\doi#1{doi:\discretionary{}{}{}#1}\fi \else
  \expandafter\ifx\csname doi\endcsname\relax
  \def\doi{doi:\discretionary{}{}{}\begingroup \urlstyle{rm}\Url}\fi \fi
\expandafter\ifx\csname selectlanguage\endcsname\relax
  \def\selectlanguage#1{}\fi

\bibitem[{Bla(2015)}]{BladderCancerStatistics2015}
Bladder cancer statistics.
\newblock https://www.cancerresearchuk.org/health-professional/cancer-statistics/statistics-by-cancer-type/bladder-cancer (2015).

\bibitem[{Van Den~Bosch and Alfred~Witjes(2011)}]{vandenboschLongtermCancerspecificSurvival2011}
Van Den~Bosch S, Alfred~Witjes J.
\newblock Long-term {{Cancer-specific Survival}} in {{Patients}} with {{High-risk}}, {{Non}}--muscle-invasive {{Bladder Cancer}} and {{Tumour Progression}}: {{A Systematic Review}}.
\newblock {\em European Urology\/} {\bf 60} (2011) 493--500.
\newblock \doi{10.1016/j.eururo.2011.05.045}.

\bibitem[{Babjuk et~al.(2017)Babjuk, B{\"o}hle, Burger, Capoun, Cohen, Comp{\'e}rat et~al.}]{babjukEAUGuidelinesNon2017}
Babjuk M, B{\"o}hle A, Burger M, Capoun O, Cohen D, Comp{\'e}rat EM, et~al.
\newblock {{EAU Guidelines}} on {{Non}}--{{Muscle-invasive Urothelial Carcinoma}} of the {{Bladder}}: {{Update}} 2016.
\newblock {\em European Urology\/} {\bf 71} (2017) 447--461.
\newblock \doi{10.1016/j.eururo.2016.05.041}.

\bibitem[{Hall et~al.(2007)Hall, Chang, Dalbagni, Pruthi, Seigne, Skinner et~al.}]{hallGuidelineManagementNonmuscle2007}
Hall MC, Chang SS, Dalbagni G, Pruthi RS, Seigne JD, Skinner EC, et~al.
\newblock Guideline for the {{Management}} of {{Nonmuscle Invasive Bladder Cancer}} ({{Stages Ta}}, {{T1}}, and {{Tis}}): 2007 {{Update}}.
\newblock {\em Journal of Urology\/} {\bf 178} (2007) 2314--2330.
\newblock \doi{10.1016/j.juro.2007.09.003}.

\bibitem[{Witjes et~al.(2014)Witjes, Comp{\'e}rat, Cowan, De~Santis, Gakis, Lebret et~al.}]{witjesEAUGuidelinesMuscleinvasive2014}
Witjes JA, Comp{\'e}rat E, Cowan NC, De~Santis M, Gakis G, Lebret T, et~al.
\newblock {{EAU Guidelines}} on {{Muscle-invasive}} and {{Metastatic Bladder Cancer}}: {{Summary}} of the 2013 {{Guidelines}}.
\newblock {\em European Urology\/} {\bf 65} (2014) 778--792.
\newblock \doi{10.1016/j.eururo.2013.11.046}.

\bibitem[{Mossanen and Gore(2014)}]{mossanenBurdenBladderCancer2014}
Mossanen M, Gore JL.
\newblock The burden of bladder cancer care: Direct and indirect costs.
\newblock {\em Current opinion in urology\/} {\bf 24} (2014) 487--491.

\bibitem[{Leal et~al.(2016)Leal, {Luengo-Fernandez}, Sullivan, and Witjes}]{lealEconomicBurdenBladder2016a}
Leal J, {Luengo-Fernandez} R, Sullivan R, Witjes JA.
\newblock Economic {{Burden}} of {{Bladder Cancer Across}} the {{European Union}}.
\newblock {\em European Urology\/} {\bf 69} (2016) 438--447.
\newblock \doi{10.1016/j.eururo.2015.10.024}.

\bibitem[{Tan et~al.(2019)Tan, Teo, Chan, Heinrich, Feber, Sarpong et~al.}]{tanMixedMethodsApproach2019}
Tan WS, Teo CH, Chan D, Heinrich M, Feber A, Sarpong R, et~al.
\newblock Mixed-methods approach to exploring patients' perspectives on the acceptability of a urinary biomarker test in replacing cystoscopy for bladder cancer surveillance.
\newblock {\em Bju International\/} {\bf 124} (2019) 408--417.
\newblock \doi{10.1111/bju.14690}.

\bibitem[{Lo et~al.(2014)Lo, Nicolle, Coffin, Gould, Maragakis, Meddings et~al.}]{loStrategiesPreventCatheterAssociated2014}
Lo E, Nicolle LE, Coffin SE, Gould C, Maragakis LL, Meddings J, et~al.
\newblock Strategies to {{Prevent Catheter-Associated Urinary Tract Infections}} in {{Acute Care Hospitals}}: 2014 {{Update}}.
\newblock {\em Infection Control \& Hospital Epidemiology\/} {\bf 35} (2014) 464--479.
\newblock \doi{10.1086/675718}.

\bibitem[{NHS(2021)}]{NHSEnglandNational2021}
{{NHS England}} {{National Cost Collection}} for the {{NHS}}.
\newblock https://www.england.nhs.uk/costing-in-the-nhs/national-cost-collection/ (2021).

\bibitem[{NHS(2020)}]{NHSEngland20202020}
{{NHS England}} 2020/21 {{National Cost Collection Data Publication}}.
\newblock https://www.england.nhs.uk/publication/2020-21-national-cost-collection-data-publication/ (2020).

\bibitem[{NHS(2019)}]{NHSEngland20192019}
{{NHS England}} 2019/20 {{National Cost Collection Data Publication}}.
\newblock https://www.england.nhs.uk/publication/2019-20-national-cost-collection-data-publication/ (2019).

\bibitem[{NHS(2018)}]{NHSEngland20182018}
{{NHS England}} 2018/19 {{National Cost Collection Data Publication}}.
\newblock https://www.england.nhs.uk/publication/2018-19-national-cost-collection-data-publication/ (2018).

\bibitem[{NHS(2015)}]{NHSReferenceCosts2015}
{{NHS}} reference costs 2015 to 2016.
\newblock https://www.gov.uk/government/publications/nhs-reference-costs-2015-to-2016 (2015).

\bibitem[{Inf(2023)}]{InflationCalculator2023}
Inflation calculator.
\newblock https://www.bankofengland.co.uk/monetary-policy/inflation/inflation-calculator (2023).

\bibitem[{Cox et~al.(2019)Cox, Saramago, Kelly, Porta, Hall, Tan et~al.}]{coxImpactsBladderCancer2019}
Cox E, Saramago P, Kelly J, Porta N, Hall E, Tan WS, et~al.
\newblock The {{Impacts}} of {{Bladder Cancer}} on {{UK Healthcare Costs}} and {{Patients}}' {{Health-Related Quality}} of {{Life}}: {{Evidence}} from the {{BOXIT Trial}} : {{Bladder}} cancer cost and health-related impacts.
\newblock {\em Clinical Genitourinary Cancer\/} {\bf 18} (2019).
\newblock \doi{10.1016/j.clgc.2019.12.004}.

\bibitem[{Kandori et~al.(2019)Kandori, Kojima, and Nishiyama}]{kandoriUpdatedPointsTNM2019}
Kandori S, Kojima T, Nishiyama H.
\newblock The updated points of {{TNM}} classification of urological cancers in the 8th edition of {{AJCC}} and {{UICC}}.
\newblock {\em Japanese Journal of Clinical Oncology\/} {\bf 49} (2019) 421--425.
\newblock \doi{10.1093/jjco/hyz017}.

\bibitem[{Edge and Compton(2010)}]{edgeAmericanJointCommittee2010}
Edge SB, Compton CC.
\newblock The {{American Joint Committee}} on {{Cancer}}: The 7th {{Edition}} of the {{AJCC Cancer Staging Manual}} and the {{Future}} of {{TNM}}.
\newblock {\em Annals of Surgical Oncology\/} {\bf 17} (2010) 1471--1474.
\newblock \doi{10.1245/s10434-010-0985-4}.

\bibitem[{Chang et~al.(2016)Chang, Boorjian, Chou, Clark, Daneshmand, Konety et~al.}]{changDiagnosisTreatmentNonMuscle2016}
Chang SS, Boorjian SA, Chou R, Clark PE, Daneshmand S, Konety BR, et~al.
\newblock Diagnosis and {{Treatment}} of {{Non-Muscle Invasive Bladder Cancer}}: {{AUA}}/{{SUO Guideline}}.
\newblock {\em Journal of Urology\/} {\bf 196} (2016) 1021--1029.
\newblock \doi{10.1016/j.juro.2016.06.049}.

\bibitem[{Hensley et~al.(2022)Hensley, Panebianco, Pietzak, Kutikov, Vikram, Galsky et~al.}]{hensleyContemporaryStagingMuscleInvasive2022}
Hensley PJ, Panebianco V, Pietzak E, Kutikov A, Vikram R, Galsky MD, et~al.
\newblock Contemporary {{Staging}} for {{Muscle-Invasive Bladder Cancer}}: {{Accuracy}} and {{Limitations}}.
\newblock {\em European Urology Oncology\/} {\bf 5} (2022) 403--411.
\newblock \doi{10.1016/j.euo.2022.04.008}.

\bibitem[{Kamat et~al.(2016)Kamat, Hahn, Efstathiou, Lerner, Malmstr{\"o}m, Choi et~al.}]{kamatBladderCancer2016}
Kamat AM, Hahn NM, Efstathiou JA, Lerner SP, Malmstr{\"o}m PU, Choi W, et~al.
\newblock Bladder cancer.
\newblock {\em The Lancet\/} {\bf 388} (2016) 2796--2810.
\newblock \doi{10.1016/S0140-6736(16)30512-8}.

\bibitem[{Clark et~al.(2013)Clark, Agarwal, Biagioli, Eisenberger, Greenberg, Herr et~al.}]{clarkClinicalPracticeGuidelines2013}
Clark PE, Agarwal N, Biagioli MC, Eisenberger MA, Greenberg RE, Herr HW, et~al.
\newblock Clinical {{Practice Guidelines}} in {{Oncology}}.
\newblock {\em Journal of the National Comprehensive Cancer Network\/} {\bf 11} (2013).

\bibitem[{Richards et~al.(2014)Richards, Smith, and Steinberg}]{richardsImportanceTransurethralResection2014}
Richards KA, Smith ND, Steinberg GD.
\newblock The {{Importance}} of {{Transurethral Resection}} of {{Bladder Tumor}} in the {{Management}} of {{Nonmuscle Invasive Bladder Cancer}}: {{A Systematic Review}} of {{Novel Technologies}}.
\newblock {\em The Journal of Urology\/} {\bf 191} (2014) 1655--1664.
\newblock \doi{10.1016/j.juro.2014.01.087}.

\bibitem[{Burger et~al.(2013)Burger, Catto, Dalbagni, Grossman, Herr, Karakiewicz et~al.}]{burgerEpidemiologyRiskFactors2013}
Burger M, Catto JW, Dalbagni G, Grossman HB, Herr H, Karakiewicz P, et~al.
\newblock Epidemiology and {{Risk Factors}} of {{Urothelial Bladder Cancer}}.
\newblock {\em European Urology\/} {\bf 63} (2013) 234--241.
\newblock \doi{10.1016/j.eururo.2012.07.033}.

\bibitem[{Chen et~al.(2005)Chen, Liou, Loh, Uang, Yu, and Shih}]{chenBladderCancerScreening2005}
Chen HI, Liou SH, Loh CH, Uang SN, Yu YC, Shih TS.
\newblock Bladder cancer screening and monitoring of 4,4{$\prime$}-{{Methylenebis}}(2-chloroaniline) exposure among workers in {{Taiwan}}.
\newblock {\em Urology\/} {\bf 66} (2005) 305--310.
\newblock \doi{10.1016/j.urology.2005.02.031}.

\bibitem[{Leta{\v s}iov{\'a} et~al.(2012)Leta{\v s}iov{\'a}, Medve{\v d}ov{\'a}, {\v S}ov{\v c}{\'i}kov{\'a}, Du{\v s}insk{\'a}, Volkovov{\'a}, Mosoiu et~al.}]{letasiovaBladderCancerReview2012}
Leta{\v s}iov{\'a} S, Medve{\v d}ov{\'a} A, {\v S}ov{\v c}{\'i}kov{\'a} A, Du{\v s}insk{\'a} M, Volkovov{\'a} K, Mosoiu C, et~al.
\newblock Bladder cancer, a review of the environmental risk factors.
\newblock {\em Environmental Health\/} {\bf 11} (2012) S11.
\newblock \doi{10.1186/1476-069X-11-S1-S11}.

\bibitem[{Cumberbatch et~al.(2018)Cumberbatch, Jubber, Black, Esperto, Figueroa, Kamat et~al.}]{cumberbatchEpidemiologyBladderCancer2018}
Cumberbatch MGK, Jubber I, Black PC, Esperto F, Figueroa JD, Kamat AM, et~al.
\newblock Epidemiology of {{Bladder Cancer}}: {{A Systematic Review}} and {{Contemporary Update}} of {{Risk Factors}} in 2018.
\newblock {\em European Urology\/} {\bf 74} (2018) 784--795.
\newblock \doi{10.1016/j.eururo.2018.09.001}.

\bibitem[{Bray et~al.(2018)Bray, Ferlay, Soerjomataram, Siegel, Torre, and Jemal}]{brayGlobalCancerStatistics2018}
Bray F, Ferlay J, Soerjomataram I, Siegel RL, Torre LA, Jemal A.
\newblock Global cancer statistics 2018: {{GLOBOCAN}} estimates of incidence and mortality worldwide for 36 cancers in 185 countries.
\newblock {\em CA: A Cancer Journal for Clinicians\/} {\bf 68} (2018) 394--424.
\newblock \doi{10.3322/caac.21492}.

\bibitem[{Farling(2017)}]{farlingBladderCancerRisk2017}
Farling KB.
\newblock Bladder cancer: {{Risk}} factors, diagnosis, and management.
\newblock {\em The Nurse Practitioner\/} {\bf 42} (2017) 26.
\newblock \doi{10.1097/01.NPR.0000512251.61454.5c}.

\bibitem[{Audenet et~al.(2018)Audenet, Attalla, and Sfakianos}]{audenetEvolutionBladderCancer2018}
Audenet F, Attalla K, Sfakianos JP.
\newblock The evolution of bladder cancer genomics: {{What}} have we learned and how can we use it?
\newblock {\em Urologic Oncology: Seminars and Original Investigations\/} {\bf 36} (2018) 313--320.
\newblock \doi{10.1016/j.urolonc.2018.02.017}.

\bibitem[{EOR(2023)}]{EORTCRiskTables2023}
{{EORTC}} risk tables: {{Predicting}} recurrence and progression in stage {{Ta T1}} bla - {{Evidencio}}.
\newblock https://www.evidencio.com/models/show/1025 (2023).

\bibitem[{Seo et~al.(2010)Seo, Kim, Park, Kim, and Chang}]{seoEfficacyEORTCScoring2010}
Seo KW, Kim BH, Park CH, Kim CI, Chang HS.
\newblock The {{Efficacy}} of the {{EORTC Scoring System}} and {{Risk Tables}} for the {{Prediction}} of {{Recurrence}} and {{Progression}} of {{Non-Muscle-Invasive Bladder Cancer}} after {{Intravesical Bacillus Calmette-Guerin Instillation}}.
\newblock {\em Korean Journal of Urology\/} {\bf 51} (2010) 165--170.
\newblock \doi{10.4111/kju.2010.51.3.165}.

\bibitem[{Pre(2023)}]{PredictingDiseaseRecurrence2023}
Predicting disease recurrence and progression - {{Uroweb}}.
\newblock https://uroweb.org/guidelines/non-muscle-invasive-bladder-cancer/chapter/predicting-disease-recurrence-and-progression (2023).

\bibitem[{Vedder et~al.(2014)Vedder, M{\'a}rquez, de~{Bekker-Grob}, Calle, Dyrskj{\o}t, Kogevinas et~al.}]{vedderRiskPredictionScores2014}
Vedder MM, M{\'a}rquez M, de~{Bekker-Grob} EW, Calle ML, Dyrskj{\o}t L, Kogevinas M, et~al.
\newblock Risk {{Prediction Scores}} for {{Recurrence}} and {{Progression}} of {{Non-Muscle Invasive Bladder Cancer}}: {{An International Validation}} in {{Primary Tumours}}.
\newblock {\em PLOS ONE\/} {\bf 9} (2014) e96849.
\newblock \doi{10.1371/journal.pone.0096849}.

\bibitem[{Krajewski et~al.(2022)Krajewski, Aumatell, Subiela, Nowak, Tukiendorf, Moschini et~al.}]{krajewskiAccuracyCUETOEORTC2022}
Krajewski W, Aumatell J, Subiela JD, Nowak {\L}, Tukiendorf A, Moschini M, et~al.
\newblock Accuracy of the {{CUETO}}, {{EORTC}} 2016 and {{EAU}} 2021 scoring models and risk stratification tables to predict outcomes in high--grade non-muscle-invasive urothelial bladder cancer.
\newblock {\em Urologic Oncology: Seminars and Original Investigations\/} {\bf 40} (2022) 491.e11--491.e19.
\newblock \doi{10.1016/j.urolonc.2022.06.008}.

\bibitem[{Fujii(2018)}]{fujiiPredictionModelsProgression2018}
Fujii Y.
\newblock Prediction models for progression of non-muscle-invasive bladder cancer: {{A}} review.
\newblock {\em International Journal of Urology\/} {\bf 25} (2018) 212--218.
\newblock \doi{10.1111/iju.13509}.

\bibitem[{Xylinas et~al.(2013)Xylinas, Kent, Kluth, Pycha, Comploj, Svatek et~al.}]{xylinasAccuracyEORTCRisk2013}
Xylinas E, Kent M, Kluth L, Pycha A, Comploj E, Svatek RS, et~al.
\newblock Accuracy of the {{EORTC}} risk tables and of the {{CUETO}} scoring model to predict outcomes in non-muscle-invasive urothelial carcinoma of the bladder.
\newblock {\em British Journal of Cancer\/} {\bf 109} (2013) 1460--1466.
\newblock \doi{10.1038/bjc.2013.372}.

\bibitem[{Born et~al.(2022)Born, Levinson, and De~Freitas}]{bornReducingHarmOveruse2022}
Born K, Levinson W, De~Freitas L.
\newblock Reducing harm from overuse of healthcare.
\newblock {\em BMJ\/}  (2022) o2787.
\newblock \doi{10.1136/bmj.o2787}.

\bibitem[{Morton(2021)}]{WarningCutsNHS2021}
Morton B.
\newblock Warning over cuts to {{NHS}} services without {\pounds}10bn extra funding.
\newblock {\em BBC News\/}  (2021).

\bibitem[{Greenfield(2018)}]{greenfieldNHSWieldsAxe2018}
Greenfield P.
\newblock {{NHS}} wields the axe on 17 'unnecessary procedures'.
\newblock {\em The Guardian\/}  (2018).

\bibitem[{Monteiro et~al.(2019)Monteiro, Witjes, Agarwal, Anderson, Bivalacqua, Bochner et~al.}]{monteiroICUDSIUInternationalConsultation2019}
Monteiro LL, Witjes JA, Agarwal PK, Anderson CB, Bivalacqua TJ, Bochner BH, et~al.
\newblock {{ICUD-SIU International Consultation}} on {{Bladder Cancer}} 2017: Management of non-muscle invasive bladder cancer.
\newblock {\em World Journal of Urology\/} {\bf 37} (2019) 51--60.
\newblock \doi{10.1007/s00345-018-2438-9}.

\bibitem[{Kim and Patel(2020)}]{kimTransurethralResectionBladder2020}
Kim LHC, Patel MI.
\newblock Transurethral resection of bladder tumour ({{TURBT}}).
\newblock {\em Translational Andrology and Urology\/} {\bf 9} (2020) 3056--3072.
\newblock \doi{10.21037/tau.2019.09.38}.

\bibitem[{Zhu et~al.(2020)Zhu, Yu, Yang, Wu, and Cheng}]{zhuTraditionalClassificationNovel2020}
Zhu S, Yu W, Yang X, Wu C, Cheng F.
\newblock Traditional {{Classification}} and {{Novel Subtyping Systems}} for {{Bladder Cancer}}.
\newblock {\em Frontiers in Oncology\/} {\bf 10} (2020).
\newblock \doi{10.3389/fonc.2020.00102}.

\bibitem[{Castaneda et~al.(2023)Castaneda, Theodorescu, Rosser, and Ahdoot}]{castanedaIdentifyingNovelBiomarkers2023}
Castaneda PR, Theodorescu D, Rosser CJ, Ahdoot M.
\newblock Identifying novel biomarkers associated with bladder cancer treatment outcomes.
\newblock {\em Frontiers in Oncology\/} {\bf 13} (2023).

\bibitem[{Zhang et~al.(2021)Zhang, Hu, Li, Ma, Othmane, Ren et~al.}]{zhangEmergingBiomarkersPredicting2021}
Zhang C, Hu J, Li H, Ma H, Othmane B, Ren W, et~al.
\newblock Emerging {{Biomarkers}} for {{Predicting Bladder Cancer Lymph Node Metastasis}}.
\newblock {\em Frontiers in Oncology\/} {\bf 11} (2021).

\bibitem[{Huang et~al.(2023)Huang, Lin, Chuang, Chuang, Pang, Wu et~al.}]{huangBodyCompositionPredictor2023}
Huang LK, Lin YC, Chuang HH, Chuang CK, Pang ST, Wu CT, et~al.
\newblock Body composition as a predictor of oncological outcome in patients with non-muscle-invasive bladder cancer receiving intravesical instillation after transurethral resection of bladder tumor.
\newblock {\em Frontiers in Oncology\/} {\bf 13} (2023) 1180888.
\newblock \doi{10.3389/fonc.2023.1180888}.

\bibitem[{Xu et~al.(2019)Xu, Wang, Du, Zhang, Li, Zhang et~al.}]{xuPredictiveNomogramIndividualized2019}
Xu X, Wang H, Du P, Zhang F, Li S, Zhang Z, et~al.
\newblock A predictive nomogram for individualized recurrence stratification of bladder cancer using multiparametric {{MRI}} and clinical risk factors.
\newblock {\em Journal of Magnetic Resonance Imaging\/} {\bf 50} (2019) 1893--1904.
\newblock \doi{10.1002/jmri.26749}.

\bibitem[{Shkolyar et~al.(2019)Shkolyar, Jia, Chang, Trivedi, Mach, Meng et~al.}]{shkolyarAugmentedBladderTumor2019}
Shkolyar E, Jia X, Chang TC, Trivedi D, Mach KE, Meng MQH, et~al.
\newblock Augmented {{Bladder Tumor Detection Using Deep Learning}}.
\newblock {\em European Urology\/} {\bf 76} (2019) 714--718.
\newblock \doi{10.1016/j.eururo.2019.08.032}.

\bibitem[{Tokuyama et~al.(2022)Tokuyama, Saito, Muraoka, Matsubara, Hashimoto, Satake et~al.}]{tokuyamaPredictionNonmuscleInvasive2022}
Tokuyama N, Saito A, Muraoka R, Matsubara S, Hashimoto T, Satake N, et~al.
\newblock Prediction of non-muscle invasive bladder cancer recurrence using machine learning of quantitative nuclear features.
\newblock {\em Modern Pathology\/} {\bf 35} (2022) 533--538.
\newblock \doi{10.1038/s41379-021-00955-y}.

\bibitem[{Pantazopoulos et~al.(1998)Pantazopoulos, Karakitsos, {Iokim-liossi}, Pouliakis, {Botsoli-stergiou}, and Dimopoulos}]{pantazopoulosBACKPROPAGATIONNEURAL1998}
Pantazopoulos D, Karakitsos P, {Iokim-liossi} A, Pouliakis A, {Botsoli-stergiou} E, Dimopoulos C.
\newblock {{BACK PROPAGATION NEURAL NETWORK IN THE DISCRIMINATION OF BENIGN FROM MALIGNANT LOWER URINARY TRACT LESIONS}}.
\newblock {\em The Journal of Urology\/} {\bf 159} (1998) 1619--1623.
\newblock \doi{10.1097/00005392-199805000-00057}.

\bibitem[{Shao et~al.(2017)Shao, Chen, Lin, Chen, Fu, Chen et~al.}]{shaoMetaboliteMarkerDiscovery2017}
Shao CH, Chen CL, Lin JY, Chen CJ, Fu SH, Chen YT, et~al.
\newblock Metabolite marker discovery for the detection of bladder cancer by comparative metabolomics.
\newblock {\em Oncotarget\/} {\bf 8} (2017) 38802--38810.
\newblock \doi{10.18632/oncotarget.16393}.

\bibitem[{Catto et~al.(2003)Catto, Linkens, Abbod, Chen, Burton, Feeley et~al.}]{cattoArtificialIntelligencePredicting2003a}
Catto JWF, Linkens DA, Abbod MF, Chen M, Burton JL, Feeley KM, et~al.
\newblock Artificial {{Intelligence}} in {{Predicting Bladder Cancer Outcome}}: {{A Comparison}} of {{Neuro-Fuzzy Modeling}} and {{Artificial Neural Networks1}}.
\newblock {\em Clinical Cancer Research\/} {\bf 9} (2003) 4172--4177.

\bibitem[{Lambin et~al.(2017)Lambin, Leijenaar, Deist, Peerlings, {de Jong}, {van Timmeren} et~al.}]{lambinRadiomicsBridgeMedical2017}
Lambin P, Leijenaar RTH, Deist TM, Peerlings J, {de Jong} EEC, {van Timmeren} J, et~al.
\newblock Radiomics: The bridge between medical imaging and personalized medicine.
\newblock {\em Nature Reviews Clinical Oncology\/} {\bf 14} (2017) 749--762.
\newblock \doi{10.1038/nrclinonc.2017.141}.

\bibitem[{Zheng et~al.(2022)Zheng, Yang, Ni, Yang, Xiong, Yan et~al.}]{zhengAccurateDiagnosisSurvival2022}
Zheng Q, Yang R, Ni X, Yang S, Xiong L, Yan D, et~al.
\newblock Accurate {{Diagnosis}} and {{Survival Prediction}} of {{Bladder Cancer Using Deep Learning}} on {{Histological Slides}}.
\newblock {\em Cancers\/} {\bf 14} (2022) 5807.
\newblock \doi{10.3390/cancers14235807}.

\bibitem[{Bychkov et~al.(2018)Bychkov, Linder, Turkki, Nordling, Kovanen, Verrill et~al.}]{bychkovDeepLearningBased2018}
Bychkov D, Linder N, Turkki R, Nordling S, Kovanen PE, Verrill C, et~al.
\newblock Deep learning based tissue analysis predicts outcome in colorectal cancer.
\newblock {\em Scientific Reports\/} {\bf 8} (2018) 3395.
\newblock \doi{10.1038/s41598-018-21758-3}.

\bibitem[{Yuan et~al.(2016)Yuan, Shi, Li, Kim, Cai, Han et~al.}]{yuanDeepGeneAdvancedCancer2016}
Yuan Y, Shi Y, Li C, Kim J, Cai W, Han Z, et~al.
\newblock {{DeepGene}}: An advanced cancer type classifier based on deep learning and somatic point mutations.
\newblock {\em BMC Bioinformatics\/} {\bf 17} (2016) 476.
\newblock \doi{10.1186/s12859-016-1334-9}.

\bibitem[{Mobadersany et~al.(2018)Mobadersany, Yousefi, Amgad, Gutman, {Barnholtz-Sloan}, Vel{\'a}zquez~Vega et~al.}]{mobadersanyPredictingCancerOutcomes2018}
Mobadersany P, Yousefi S, Amgad M, Gutman DA, {Barnholtz-Sloan} JS, Vel{\'a}zquez~Vega JE, et~al.
\newblock Predicting cancer outcomes from histology and genomics using convolutional networks.
\newblock {\em Proceedings of the National Academy of Sciences\/} {\bf 115} (2018) E2970--E2979.
\newblock \doi{10.1073/pnas.1717139115}.

\bibitem[{Cha et~al.(2016)Cha, Hadjiiski, Samala, Chan, Cohan, Caoili et~al.}]{chaBladderCancerSegmentation2016}
Cha KH, Hadjiiski LM, Samala RK, Chan HP, Cohan RH, Caoili EM, et~al.
\newblock Bladder {{Cancer Segmentation}} in {{CT}} for {{Treatment Response Assessment}}: {{Application}} of {{Deep-Learning Convolution Neural Network}}---{{A Pilot Study}}.
\newblock {\em Tomography\/} {\bf 2} (2016) 421--429.
\newblock \doi{10.18383/j.tom.2016.00184}.

\bibitem[{Jansen et~al.(2020)Jansen, Lucas, Bosschieter, {de Boer}, Meijer, {van Leeuwen} et~al.}]{jansenAutomatedDetectionGrading2020}
Jansen I, Lucas M, Bosschieter J, {de Boer} OJ, Meijer SL, {van Leeuwen} TG, et~al.
\newblock Automated {{Detection}} and {{Grading}} of {{Non}}--{{Muscle-Invasive Urothelial Cell Carcinoma}} of the {{Bladder}}.
\newblock {\em The American Journal of Pathology\/} {\bf 190} (2020) 1483--1490.
\newblock \doi{10.1016/j.ajpath.2020.03.013}.

\bibitem[{Wang et~al.(2019)Wang, Hu, Yao, Chen, Li, Chen et~al.}]{wangRadiomicsAnalysisMultiparametric2019}
Wang H, Hu D, Yao H, Chen M, Li S, Chen H, et~al.
\newblock Radiomics analysis of multiparametric {{MRI}} for the preoperative evaluation of pathological grade in bladder cancer tumors.
\newblock {\em European Radiology\/} {\bf 29} (2019) 6182--6190.
\newblock \doi{10.1007/s00330-019-06222-8}.

\bibitem[{Mucaki et~al.(2019)Mucaki, Zhao, Lizotte, and Rogan}]{mucakiPredictingResponsesPlatin2019a}
Mucaki EJ, Zhao JZL, Lizotte DJ, Rogan PK.
\newblock Predicting responses to platin chemotherapy agents with biochemically-inspired machine learning.
\newblock {\em Signal Transduction and Targeted Therapy\/} {\bf 4} (2019) 1.
\newblock \doi{10.1038/s41392-018-0034-5}.

\bibitem[{Liberati et~al.(2009)Liberati, Altman, Tetzlaff, Mulrow, G{\o}tzsche, Ioannidis et~al.}]{liberatiPRISMAStatementReporting2009}
Liberati A, Altman DG, Tetzlaff J, Mulrow C, G{\o}tzsche PC, Ioannidis JPA, et~al.
\newblock The {{PRISMA Statement}} for {{Reporting Systematic Reviews}} and {{Meta-Analyses}} of {{Studies That Evaluate Health Care Interventions}}: {{Explanation}} and {{Elaboration}}.
\newblock {\em PLoS Medicine\/} {\bf 6} (2009) e1000100.
\newblock \doi{10.1371/journal.pmed.1000100}.

\bibitem[{Chen et~al.(2021)Chen, Jiang, Zheng, Shao, Wang, Zhang et~al.}]{chenClinicalUseMachine2021}
Chen S, Jiang L, Zheng X, Shao J, Wang T, Zhang E, et~al.
\newblock Clinical use of machine learning-based pathomics signature for diagnosis and survival prediction of bladder cancer.
\newblock {\em Cancer Science\/} {\bf 112} (2021) 2905--2914.
\newblock \doi{10.1111/cas.14927}.

\bibitem[{Drachneris et~al.(2023)Drachneris, Rasmusson, Morkunas, Fabijonavicius, Cekauskas, Jankevicius et~al.}]{drachnerisCD8CellDensity2023a}
Drachneris J, Rasmusson A, Morkunas M, Fabijonavicius M, Cekauskas A, Jankevicius F, et~al.
\newblock {{CD8}}+ {{Cell Density Gradient}} across the {{Tumor Epithelium}}--{{Stromal Interface}} of {{Non-Muscle Invasive Papillary Urothelial Carcinoma Predicts Recurrence-Free Survival}} after {{BCG Immunotherapy}}.
\newblock {\em Cancers\/} {\bf 15} (2023) 1205.
\newblock \doi{10.3390/cancers15041205}.

\bibitem[{Levy et~al.(2023)Levy, Chan, Marotti, Rodrigues, Ismail, Kerr et~al.}]{levyExaminingLongitudinalMarkers2023}
Levy JJ, Chan N, Marotti JD, Rodrigues NJ, Ismail AAO, Kerr DA, et~al.
\newblock Examining longitudinal markers of bladder cancer recurrence through a semiautonomous machine learning system for quantifying specimen atypia from urine cytology.
\newblock {\em Cancer Cytopathology\/} {\bf 131} (2023) 561--573.
\newblock \doi{10.1002/cncy.22725}.

\bibitem[{Wang et~al.(2024)Wang, Zhu, Wang, Qin, Wang, Liu et~al.}]{wangPredictionNonmuscleInvasive2024}
Wang GY, Zhu JF, Wang QC, Qin JX, Wang XL, Liu X, et~al.
\newblock Prediction of non-muscle invasive bladder cancer recurrence using deep learning of pathology image.
\newblock {\em Scientific Reports\/} {\bf 14} (2024) 18931.
\newblock \doi{10.1038/s41598-024-66870-9}.

\bibitem[{Huang et~al.(2024)Huang, Huang, Kaggie, Cai, Yang, Wei et~al.}]{huangMultiparametricMRIBased2024}
Huang H, Huang Y, Kaggie JD, Cai Q, Yang P, Wei J, et~al.
\newblock Multiparametric {{{\textsc{MRI}}}} -{{Based Deep Learning Radiomics Model}} for {{Assessing}} 5-{{Year Recurrence Risk}} in {{Non}}-{{Muscle Invasive Bladder Cancer}}.
\newblock {\em Journal of Magnetic Resonance Imaging\/}  (2024) jmri.29574.
\newblock \doi{10.1002/jmri.29574}.

\bibitem[{Wang et~al.(2023)Wang, Zhang, Miao, Hou, Chen, Huang et~al.}]{wangDeepLearningSignature2023}
Wang H, Zhang M, Miao J, Hou F, Chen Y, Huang Y, et~al.
\newblock Deep learning signature based on multiphase enhanced {{CT}} for bladder cancer recurrence prediction: A multi-center study.
\newblock {\em eClinicalMedicine\/} {\bf 66} (2023) 102352.
\newblock \doi{10.1016/j.eclinm.2023.102352}.

\bibitem[{Lee et~al.(2021)Lee, Choo, Yoo, Cho, Son, and Jeong}]{leeIntravesicalProstaticProtrusion2021}
Lee J, Choo MS, Yoo S, Cho MC, Son H, Jeong H.
\newblock Intravesical {{Prostatic Protrusion}} and {{Prognosis}} of {{Non-Muscle Invasive Bladder Cancer}}: {{Analysis}} of {{Long-Term Data}} over 5 {{Years}} with {{Machine-Learning Algorithms}}.
\newblock {\em Journal of Clinical Medicine\/} {\bf 10} (2021) 4263.
\newblock \doi{10.3390/jcm10184263}.

\bibitem[{Zhao et~al.(2022)Zhao, Chen, Fang, Su, Xu, Wang et~al.}]{zhaoPredictionPrognosisRecurrence2022}
Zhao H, Chen Z, Fang Y, Su M, Xu Y, Wang Z, et~al.
\newblock Prediction of {{Prognosis}} and {{Recurrence}} of {{Bladder Cancer}} by {{ECM-Related Genes}}.
\newblock {\em Journal of Immunology Research\/} {\bf 2022} (2022) 1793005.
\newblock \doi{10.1155/2022/1793005}.

\bibitem[{Cai et~al.(2010)Cai, Nesi, Canto, Mondaini, {Mauro Piazzini}, and Bartoletti}]{caiPrognosticRoleLoss2010}
Cai T, Nesi G, Canto MD, Mondaini N, {Mauro Piazzini}, Bartoletti R.
\newblock Prognostic {{Role}} of {{Loss}} of {{Heterozygosity}} on {{Chromosome}} 18 in {{Patients With Low-Risk Nonmuscle-Invasive Bladder Cancer}}: {{Results}} from a {{Prospective Study}}.
\newblock {\em Journal of Surgical Research\/} {\bf 161} (2010) 89--94.
\newblock \doi{10.1016/j.jss.2008.10.017}.

\bibitem[{Sikic et~al.(2021)Sikic, Taubert, Breyer, Eckstein, Weyerer, Keck et~al.}]{sikicPrognosticValueFGFR32021a}
Sikic D, Taubert H, Breyer J, Eckstein M, Weyerer V, Keck B, et~al.
\newblock The {{Prognostic Value}} of {{FGFR3 Expression}} in {{Patients}} with {{T1 Non-Muscle Invasive Bladder Cancer}}.
\newblock {\em Cancer Management and Research\/} {\bf 13} (2021) 6567--6578.
\newblock \doi{10.2147/CMAR.S318893}.

\bibitem[{Bertz et~al.(2014)Bertz, Otto, Denzinger, Wieland, Burger, St{\"o}hr et~al.}]{bertzCombinationCK20Ki672014}
Bertz S, Otto W, Denzinger S, Wieland WF, Burger M, St{\"o}hr R, et~al.
\newblock Combination of {{CK20}} and {{Ki-67 Immunostaining Analysis Predicts Recurrence}}, {{Progression}}, and {{Cancer-Specific Survival}} in {{pT1 Urothelial Bladder Cancer}}.
\newblock {\em European Urology\/} {\bf 65} (2014) 218--226.
\newblock \doi{10.1016/j.eururo.2012.05.033}.

\bibitem[{Wang et~al.(2017)Wang, Que, Suo, Han, Tao, Huang et~al.}]{wangEvaluationNMP22BladderChek2017}
Wang Z, Que H, Suo C, Han Z, Tao J, Huang Z, et~al.
\newblock Evaluation of the {{NMP22 BladderChek}} test for detecting bladder cancer: A systematic review and meta-analysis.
\newblock {\em Oncotarget\/} {\bf 8} (2017) 100648--100656.
\newblock \doi{10.18632/oncotarget.22065}.

\bibitem[{Ponsky et~al.(2001)Ponsky, Sharma, Pandrangi, Kedia, Nelson, Agarwal et~al.}]{ponskySCREENINGMONITORINGBLADDER2001}
Ponsky LE, Sharma {\relax SHASHIKALA}, Pandrangi {\relax LAKSHMI}, Kedia {\relax SUMITA}, Nelson {\relax DAVID}, Agarwal {\relax ASHOK}, et~al.
\newblock {{SCREENING AND MONITORING FOR BLADDER CANCER}}: {{REFINING THE USE OF NMP22}}.
\newblock {\em The Journal of Urology\/} {\bf 166} (2001) 75--78.
\newblock \doi{10.1016/S0022-5347(05)66080-6}.

\bibitem[{Urbanowicz et~al.(2013)Urbanowicz, Andrew, Karagas, and Moore}]{urbanowiczRoleGeneticHeterogeneity2013a}
Urbanowicz RJ, Andrew AS, Karagas MR, Moore JH.
\newblock Role of genetic heterogeneity and epistasis in bladder cancer susceptibility and outcome: A learning classifier system approach.
\newblock {\em Journal of the American Medical Informatics Association\/} {\bf 20} (2013) 603--612.
\newblock \doi{10.1136/amiajnl-2012-001574}.

\bibitem[{Wu et~al.(2009)Wu, Hildebrandt, and Chang}]{wuGenomewideAssociationStudies2009}
Wu X, Hildebrandt MAT, Chang DW.
\newblock Genome-wide association studies of bladder cancer risk: A field synopsis of progress and potential applications.
\newblock {\em Cancer and Metastasis Reviews\/} {\bf 28} (2009) 269--280.
\newblock \doi{10.1007/s10555-009-9190-y}.

\bibitem[{Menon and Rosand(2021)}]{menonFindingPlaceCandidate2021}
Menon DK, Rosand J.
\newblock Finding a {{Place}} for {{Candidate Gene Studies}} in a {{Genome-Wide Association Study World}}.
\newblock {\em JAMA Network Open\/} {\bf 4} (2021) e2118594.
\newblock \doi{10.1001/jamanetworkopen.2021.18594}.

\bibitem[{{Garcia-Closas} et~al.(2011){Garcia-Closas}, Ye, Rothman, Figueroa, Malats, Dinney et~al.}]{garcia-closasGenomewideAssociationStudy2011}
{Garcia-Closas} M, Ye Y, Rothman N, Figueroa JD, Malats N, Dinney CP, et~al.
\newblock A genome-wide association study of bladder cancer identifies a new susceptibility locus within {{SLC14A1}}, a urea transporter gene on chromosome 18q12.3.
\newblock {\em Human Molecular Genetics\/} {\bf 20} (2011) 4282--4289.
\newblock \doi{10.1093/hmg/ddr342}.

\bibitem[{Rafnar et~al.(2014)Rafnar, Sulem, Thorleifsson, Vermeulen, Helgason, Saemundsdottir et~al.}]{rafnarGenomewideAssociationStudy2014}
Rafnar T, Sulem P, Thorleifsson G, Vermeulen SH, Helgason H, Saemundsdottir J, et~al.
\newblock Genome-wide association study yields variants at 20p12.2 that associate with urinary bladder cancer.
\newblock {\em Human Molecular Genetics\/} {\bf 23} (2014) 5545--5557.
\newblock \doi{10.1093/hmg/ddu264}.

\bibitem[{Wang et~al.(2016)Wang, Li, Chu, Lv, Ye, Ding et~al.}]{wangGenomeWideAssociationStudy2016}
Wang M, Li Z, Chu H, Lv Q, Ye D, Ding Q, et~al.
\newblock Genome-{{Wide Association Study}} of {{Bladder Cancer}} in a {{Chinese Cohort Reveals}} a {{New Susceptibility Locus}} at 5q12.3.
\newblock {\em Cancer Research\/} {\bf 76} (2016) 3277--3284.
\newblock \doi{10.1158/0008-5472.CAN-15-2564}.

\bibitem[{{L{\'o}pez de Maturana} et~al.(2016){L{\'o}pez de Maturana}, Picornell, {Masson-Lecomte}, Kogevinas, M{\'a}rquez, Carrato et~al.}]{lopezdematuranaPredictionNonmuscleInvasive2016}
{L{\'o}pez de Maturana} E, Picornell A, {Masson-Lecomte} A, Kogevinas M, M{\'a}rquez M, Carrato A, et~al.
\newblock Prediction of non-muscle invasive bladder cancer outcomes assessed by innovative multimarker prognostic models.
\newblock {\em BMC Cancer\/} {\bf 16} (2016) 351.
\newblock \doi{10.1186/s12885-016-2361-7}.

\bibitem[{Chang et~al.(2024)Chang, Chen, Yin, Wang, Dai, Wu et~al.}]{changComprehensiveUrinaryProteome2024}
Chang Q, Chen Y, Yin J, Wang T, Dai Y, Wu Z, et~al.
\newblock Comprehensive {{Urinary Proteome Profiling Analysis Identifies Diagnosis}} and {{Relapse Surveillance Biomarkers}} for {{Bladder Cancer}}.
\newblock {\em Journal of Proteome Research\/} {\bf 23} (2024) 2241--2252.
\newblock \doi{10.1021/acs.jproteome.4c00199}.

\bibitem[{Frantzi et~al.(2016)Frantzi, Van~Kessel, Zwarthoff, Marquez, Rava, Malats et~al.}]{frantziDevelopmentValidationUrinebased2016a}
Frantzi M, Van~Kessel KE, Zwarthoff EC, Marquez M, Rava M, Malats N, et~al.
\newblock Development and {{Validation}} of {{Urine-based Peptide Biomarker Panels}} for {{Detecting Bladder Cancer}} in a {{Multi-center Study}}.
\newblock {\em Clinical Cancer Research\/} {\bf 22} (2016) 4077--4086.
\newblock \doi{10.1158/1078-0432.CCR-15-2715}.

\bibitem[{Krochmal et~al.(2019)Krochmal, {van Kessel}, Zwarthoff, Belczacka, Pejchinovski, Vlahou et~al.}]{krochmalUrinaryPeptidePanel2019}
Krochmal M, {van Kessel} KEM, Zwarthoff EC, Belczacka I, Pejchinovski M, Vlahou A, et~al.
\newblock Urinary peptide panel for prognostic assessment of bladder cancer relapse.
\newblock {\em Scientific Reports\/} {\bf 9} (2019) 7635.
\newblock \doi{10.1038/s41598-019-44129-y}.

\bibitem[{Zhan et~al.(2018)Zhan, Du, Wang, Jiang, Zhang, Li et~al.}]{zhanExpressionSignaturesExosomal2018a}
Zhan Y, Du L, Wang L, Jiang X, Zhang S, Li J, et~al.
\newblock Expression signatures of exosomal long non-coding {{RNAs}} in urine serve as novel non-invasive biomarkers for diagnosis and recurrence prediction of bladder cancer.
\newblock {\em Molecular Cancer\/} {\bf 17} (2018) 142.
\newblock \doi{10.1186/s12943-018-0893-y}.

\bibitem[{Gogalic et~al.(2017)Gogalic, Sauer, Doppler, Heinzel, Perco, Lukas et~al.}]{gogalicValidationProteinPanel2017}
Gogalic S, Sauer U, Doppler S, Heinzel A, Perco P, Lukas A, et~al.
\newblock Validation of a protein panel for the noninvasive detection of recurrent non-muscle invasive bladder cancer.
\newblock {\em Biomarkers\/} {\bf 22} (2017) 674--681.
\newblock \doi{10.1080/1354750X.2016.1276628}.

\bibitem[{Ajili(2014)}]{ajiliPrognosticValueArtificial2014}
Ajili F.
\newblock Prognostic {{Value}} of {{Artificial Neural Network}} in {{Predicting Bladder Cancer Recurrence}} after {{BCG Immunotherapy}}.
\newblock {\em Journal of Cytology \& Histology\/} {\bf 05} (2014).
\newblock \doi{10.4172/2157-7099.1000226}.

\bibitem[{Zhang and Ma(2024)}]{zhangPredictiveValueTotal2024}
Zhang X, Ma L.
\newblock Predictive {{Value}} of the {{Total Bilirubin}} and {{CA50 Screened Based}} on {{Machine Learning}} for {{Recurrence}} of {{Bladder Cancer Patients}}.
\newblock {\em Cancer Management and Research\/} {\bf 16} (2024) 537--546.
\newblock \doi{10.2147/CMAR.S457269}.

\bibitem[{Schwarz et~al.(2024)Schwarz, Sobania, and Rothlauf}]{schwarzRelevantFeaturesRecurrence2024}
Schwarz L, Sobania D, Rothlauf F.
\newblock On relevant features for the recurrence prediction of urothelial carcinoma of the bladder.
\newblock {\em International Journal of Medical Informatics\/} {\bf 186} (2024) 105414.
\newblock \doi{10.1016/j.ijmedinf.2024.105414}.

\bibitem[{Lucas et~al.(2022)Lucas, Jansen, {van Leeuwen}, Oddens, {de Bruin}, and Marquering}]{lucasDeepLearningBased2022a}
Lucas M, Jansen I, {van Leeuwen} TG, Oddens JR, {de Bruin} DM, Marquering HA.
\newblock Deep {{Learning}}--based {{Recurrence Prediction}} in {{Patients}} with {{Non}}--muscle-invasive {{Bladder Cancer}}.
\newblock {\em European Urology Focus\/} {\bf 8} (2022) 165--172.
\newblock \doi{10.1016/j.euf.2020.12.008}.

\bibitem[{Jobczyk et~al.(2022)Jobczyk, Stawiski, Kaszkowiak, Rajwa, R{\'o}{\.z}a{\'n}ski, Soria et~al.}]{jobczykDeepLearningbasedRecalibration2022}
Jobczyk M, Stawiski K, Kaszkowiak M, Rajwa P, R{\'o}{\.z}a{\'n}ski W, Soria F, et~al.
\newblock Deep {{Learning-based Recalibration}} of the {{CUETO}} and {{EORTC Prediction Tools}} for {{Recurrence}} and {{Progression}} of {{Non}}--muscle-invasive {{Bladder Cancer}}.
\newblock {\em European Urology Oncology\/} {\bf 5} (2022) 109--112.
\newblock \doi{10.1016/j.euo.2021.05.006}.

\bibitem[{Hasnain et~al.(2019)Hasnain, Mason, Gill, Miranda, Gill, Kuhn et~al.}]{hasnainMachineLearningModels2019}
Hasnain Z, Mason J, Gill K, Miranda G, Gill IS, Kuhn P, et~al.
\newblock Machine learning models for predicting post-cystectomy recurrence and survival in bladder cancer patients.
\newblock {\em PLOS ONE\/} {\bf 14} (2019) e0210976.
\newblock \doi{10.1371/journal.pone.0210976}.

\end{thebibliography}
}

\end{document}